\documentclass[10pt,final,journal]{IEEEtran}
\usepackage[noadjust]{cite}  
\usepackage{graphicx}
\usepackage{amsmath}

\usepackage{hyperref}      
\usepackage{url}               
\usepackage{amsfonts}     
\usepackage{nicefrac}       
\usepackage{microtype}    
\usepackage{lipsum}
\usepackage{bm, color, soul}
\usepackage[caption=false,font=footnotesize]{subfig} 

\usepackage{tabularx, booktabs}       
\usepackage[para, flushleft]{threeparttable}
\newcolumntype{b}{>{\centering\arraybackslash}X}
\newcolumntype{B}{>{\hsize=1.25\hsize}X}
\newcolumntype{m}{>{\hsize=.5\hsize}b}
\newcolumntype{s}{>{\hsize=.25\hsize}b}

\begin{document}

\title{Satellite Pose Estimation Challenge: Dataset, Competition Design and Results}%

\author{Mate~Kisantal,
    Sumant~Sharma,~\IEEEmembership{Member,~IEEE,}
    Tae~Ha~Park,~\IEEEmembership{Student Member,~IEEE,}
    Dario~Izzo,
    Marcus~M\"{a}rtens,
    and~Simone~D'Amico%
    \thanks{M.~Kisantal was with the Advanced Concepts Team of the European Space Agency, Noordwijk, The Netherlands (email: kisantal.mate@gmail.com)}%
    \thanks{S.~Sharma is a Computer Vision Engineer at Wisk Aero LLC, 2700 Broderick Way, Mountain View, CA 94043 USA (e-mail: contact@sumantsharma.net)}%
    \thanks{T.~Park, and S.~D'Amico are with the Department of Aeronautics and Astronautics, Stanford University, 496 Lomita Mall, Stanford, CA 94305 USA (email: \{tpark94, damicos\}@stanford.edu).}%
    \thanks{D.~Izzo and M.~M\"{a}rtens are with the Advanced Concepts Team of the European Space Agency, Noordwijk, The Netherlands (email: \{dario.izzo, marcus.maertens\}@esa.int)}
	\thanks{\copyright~2020 IEEE. Personal use of this material is permitted.  Permission from IEEE must be obtained for all other uses, in any current or future media, including reprinting/republishing this material for advertising or promotional purposes, creating new collective works, for resale or redistribution to servers or lists, or reuse of any copyrighted component of this work in other works.}
}


\markboth{IEEE Transactions on Aerospace and Electronic Systems}%
{Kisantal \MakeLowercase{\textit{et al.}}: Satellite Pose Estimation Challenge: Dataset, Competition Design, and Results}
%

\IEEEpubid{0000--0000/00\$00.00~\copyright~2020 IEEE}
\IEEEpubidadjcol

\maketitle

\begin{abstract}
Reliable pose estimation of uncooperative satellites is a key technology for enabling future on-orbit servicing and debris removal missions. The Kelvins Satellite Pose Estimation Challenge aims at evaluating and comparing monocular vision-based approaches and pushing the state-of-the-art on this problem. This work is based on the Satellite Pose Estimation Dataset, the first publicly available machine learning set of synthetic and real spacecraft imageries. The choice of dataset reflects one of the unique challenges associated with spaceborne computer vision tasks, namely the lack of spaceborne images to train and validate the developed algorithms. This work briefly reviews the basic properties and the collection process of the dataset which was made publicly available. The competition design, including the definition of performance metrics and the adopted testbed, is also discussed. The main contribution of this paper is the analysis of the submissions of the 48 competitors, which compares the performance of different approaches and uncovers what factors make the satellite pose estimation problem especially challenging.
\end{abstract}



\section{Introduction}
\label{sec:introduction} 

\IEEEPARstart{I}{n} recent years, mission concepts such as debris removal and on-orbit servicing have gained increasing attention from academia and industry in order to address the congestion in Earth orbits and extend the lifetime of geostationary satellites. These include the RemoveDEBRIS mission by Surrey Space Centre \cite{removedebris}, the Phoenix program by DARPA \cite{phoenix_darpa}, the Restore-L mission by NASA \cite{restore_L}, and the on-orbit servicing programs proposed by startup companies such as Infinite Orbits\footnote{https://www.infiniteorbits.io/} and Effective Space\footnote{https://www.effective.space/}. A key to performing these tasks is the availability of the target spacecraft's position and attitude relative to the servicer spacecraft (i.e.,~pose). However, the targets of interest, including defunct satellites and debris pieces, are uncooperative and thus incapable of providing the servicer the information on their state. Moreover, the servicer cannot rely on the availability of known fiduciary markers on these targets. Overall, the servicer must be able to estimate and predict the target's relative pose on-board without human-in-the-loop. It is especially attractive to perform pose estimation using a vision-based sensor such as a camera due to its small mass and power requirements compared to other active sensors such as Light Detection and Ranging (LIDAR) or Range Detection and Ranging (RADAR). Moreover, monocular cameras are favored over stereo systems due to their relative simplicity and the fact that spacecraft, especially emerging small spacecraft such as CubeSats, do not allow for a large enough baseline to make stereovision effective. In order to enable autonomous pose estimation, the servicer then must harness fast and robust computer vision algorithms to compute relative position and attitude of the target from a single or a set of monocular images. Cassinis et al.~\cite{Cassinis2019Review} provide a comprehensive survey of different approaches for pose estimation of uncooperative spacecraft.

\IEEEpubidadjcol

Starting with the the success of AlexNet \cite{Krizhevsky2012} in the ILSVRC challenge \cite{ilsvrc} in 2012, deep learning models have been outperforming traditional approaches on a number of computer vision problems. However, deep learning relies on large annotated datasets. While there is a plethora of large-scale datasets for various terrestrial applications of computer vision and pose estimation that allows training the state-of-the-art machine learning models, there is a lack of such datasets for spacecraft pose estimation. The main reason arises from the difficulty of acquiring thousands of spaceborne images of the desired target spacecraft with accurately annotated pose labels. Moreover, a lack of common datasets makes it impossible to systematically evaluate and compare the performance of different pose estimation algorithms. In order to address these difficulties, the Satellite Pose Estimation Challenge (SPEC) was organized by the Space Rendezvous Laboratory (SLAB) at Stanford University and the Advanced Concepts Team (ACT) of the European Space Agency (ESA). The challenge was hosted on the ACT's Kelvins competition website\footnote{https://kelvins.esa.int/}, a platform hosting a number of space-related competitions. The primary aim of the SPEC was to provide a common benchmark for satellite pose estimation algorithms, identify the state-of-the-art, and show where further improvements can be made. Furthermore, such dedicated challenges have potential to raise awareness of the problems of the satellite pose estimation in the wider scientific community, bringing in new ideas and researchers to this field.

The dataset for the SPEC, named Spacecraft Pose Estimation Dataset (SPEED) \cite{Sharma2019, SPEEDdoi}, mostly consists of synthetic images and the submissions were solely ranked by their accuracy as evaluated on these images. The dataset also includes a smaller amount of real images which were collected using a realistic satellite mockup and the Testbed for Rendezvous and Optical Navigation (TRON) facility of SLAB \cite{sharmaThesis2019, BeierleThesis2019}. Even though the domain adaptation was not the main focus of the competition, evaluating the submissions on these images provides an indication of the generalization capability of the proposed algorithms.

The main contribution of this work is the analysis of the SPEC results. On the one hand, samples of the dataset are ranked based on performance of the submitted algorithms to uncover which factors contribute to the difficulty of the pose estimation task the most. Target distance and background were found to be the main challenges. On the other hand, an analysis of the submissions and comparison of the efficacy of different approaches are presented based on a survey conducted among the participants. Perspective-n-Point (PnP) solver-based approaches were found to be significantly more accurate compared to direct pose estimation approaches. Including a separate detection step was also found to be an important element of high performing pose estimation pipelines. It allows cropping the relevant part of the images and zooming on the satellite, which brings significant benefits in terms of orientation accuracy.

After a review of the related pose estimation research in Section \ref{sec:related}, Section \ref{sec:dataset} discusses the creation of the dataset, and Section \ref{sec:competition_design} briefly discusses the competition design considerations. This is followed by an in-depth analysis of the final submissions in Section \ref{sec:competition_results}. Finally, the recommendations for further improvements are given in Section \ref{sec:conclusion}.

\section{Related Work}
\label{sec:related}

The classical approach to monocular-based pose estimation of a target spacecraft \cite{Cropp2002PoseEO, Leinz2008_OrbitalExpress, Zhang2005_pose, Petit2011_CaseStudy, grompone2015_phdthesis, damico_benn_jorgensen_2014, kanani2012} would first extract hand-crafted features of the target from a 2D image. These features include Harris corners \cite{Harris88acombined}, Canny edges \cite{Canny:1986:CAE:11274.11275}, lines via Hough transform \cite{ballard_1981}, or scale-invariant features such as SIFT \cite{Lowe2004}, SURF \cite{Bay:2008:SRF:1370312.1370556}, and ORB features \cite{Rublee:2011:OEA:2355573.2356268}. Upon successful extraction of said features, iterative algorithms are required to predict the best pose solution that minimizes a certain error criterion in the presence of outliers and unknown features correspondences. The process is crucial in providing a good initial pose estimate to the on-board vision-based navigation system \cite{sharma_damico_2017, KimJunkins2007_KFPose}. Earlier works on initial pose estimation tended to rely on a coarse a priori knowledge of the target's pose \cite{Cropp2002PoseEO, Zhang2005_pose, Petit2011_CaseStudy} or assumed the availability of active fiduciary markers or sensors on the target \cite{Leinz2008_OrbitalExpress}. Without making any such assumptions, D'Amico et al.~\cite{damico_benn_jorgensen_2014} were one of the first to publish pose estimation results using Hough transform and Canny edge detector on spaceborne images captured during the rendezvous phase of the PRISMA mission \cite{damico_benn_jorgensen_2014, PRISMA_chapter}. By grouping edge features into a geometrically meaningful shape, they were able to reduce the size of the feature correspondence search space. The work was followed by Sharma et al.~\cite{Sharma2018_RobustPoseInitial} who additionally introduced Weak Gradient Elimination (WGE) technique to essentially separate the spacecraft's edge features from the weak edge features of the background. While the proposed architecture showed improved performance on the spaceborne images from the PRISMA mission, the method was affected by low availability of high confidence solutions.

On the other hand, recent years have seen a significant breakthrough in computer vision with the advent of Deep Neural Networks (DNN). It was made possible by increasing computational resources represented by the Graphical Processing Units (GPU) and the availability of large-scale datasets to train the DNN, such as ImageNet for classification \cite{Krizhevsky2012}, MS COCO for object detection \cite{Lin2014COCO}, and LINEMOD for pose estimation \cite{Hinterstoisser2013_LINEMOD} of ordinary household objects. While various DNN-based approaches have been proposed to perform pose estimation \cite{Tulsiani2015, Su2015_RenderForCNN, Kendall2015_PoseNet, Rad2017_BB8, Kehl2017_ssd6d, Sundermeyer_2018_ECCV, Mahendran2017, Xiang2018_PoseCNN, Tekin2018, Zhao2018_KPD, Peng2019_PVNet}, current state-of-the-art methods employ Convolutional Neural Networks (CNN) that either directly predict the 6D pose or an intermediate information that can be used to compute the 6D pose, notably a set of keypoints defined \emph{a priori}. For example, PoseCNN \cite{Xiang2018_PoseCNN} directly regresses a 3D translation vector and a unit quaternion representing the relative attitude of the target, whereas SPN \cite{Sharma2019, sharmaThesis2019} poses attitude prediction as a classification problem by discretizing the viewpoint space into a finite number of bins. Most recently, architectures like KPD \cite{Zhao2018_KPD} and PVNet \cite{Peng2019_PVNet} have been proposed to predict the locations of the 2D keypoints on the target's surface. Given the corresponding 3D coordinates of the keypoints from available models, one can solve the PnP problem \cite{Lepetit2008} to compute the relative position and attitude. It is noteworthy to mention that terrestrial applications of the object pose estimation are not typically subject to strict navigation and computation requirements as for satellite on-orbit servicing.

\section{Dataset}
\label{sec:dataset}

This section provides a high-level description of SPEED, which comprises the training and test images of this challenge. SPEED \cite{SPEEDdoi} represents the first publicly available machine learning data set for spacecraft pose estimation initially released in February 2019. The images of the Tango spacecraft from the PRISMA mission \cite{damico_benn_jorgensen_2014, PRISMA_chapter} are generated from two different sources, referred to as \textit{synthetic} and \textit{real} images in the following. Both images are created using the same camera model. Specifically, the real images are captured using the Point Grey Grasshopper 3 camera with a Xenoplan 1.9/17 mm lens, while the synthetic images are created using the same camera properties. The ground-truth pose labels, consisting of the translation vector and a unit quaternion describing the relative orientation of the Tango spacecraft with respect to the camera, are released along with the associated training images. The readers are encouraged to read \cite{Sharma2019} and \cite{sharmaThesis2019} for more details on the dataset. 

\begin{figure*}[!t]
	\centering
	\includegraphics[width=1\textwidth]{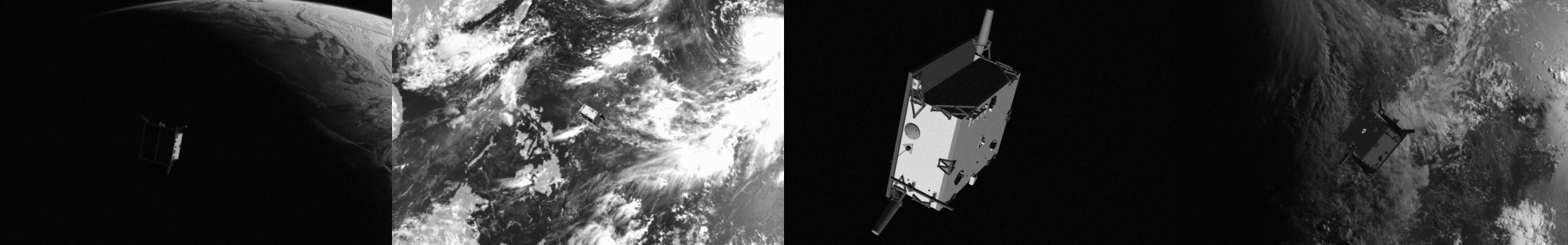}
	\caption{Examples of synthetic training images from SPEED.}
	\label{fig:synthetic_examples}
\end{figure*}

\begin{figure*}[!t]
	\centering
	\subfloat[Flight imagery]{\includegraphics[width=0.16\textwidth]{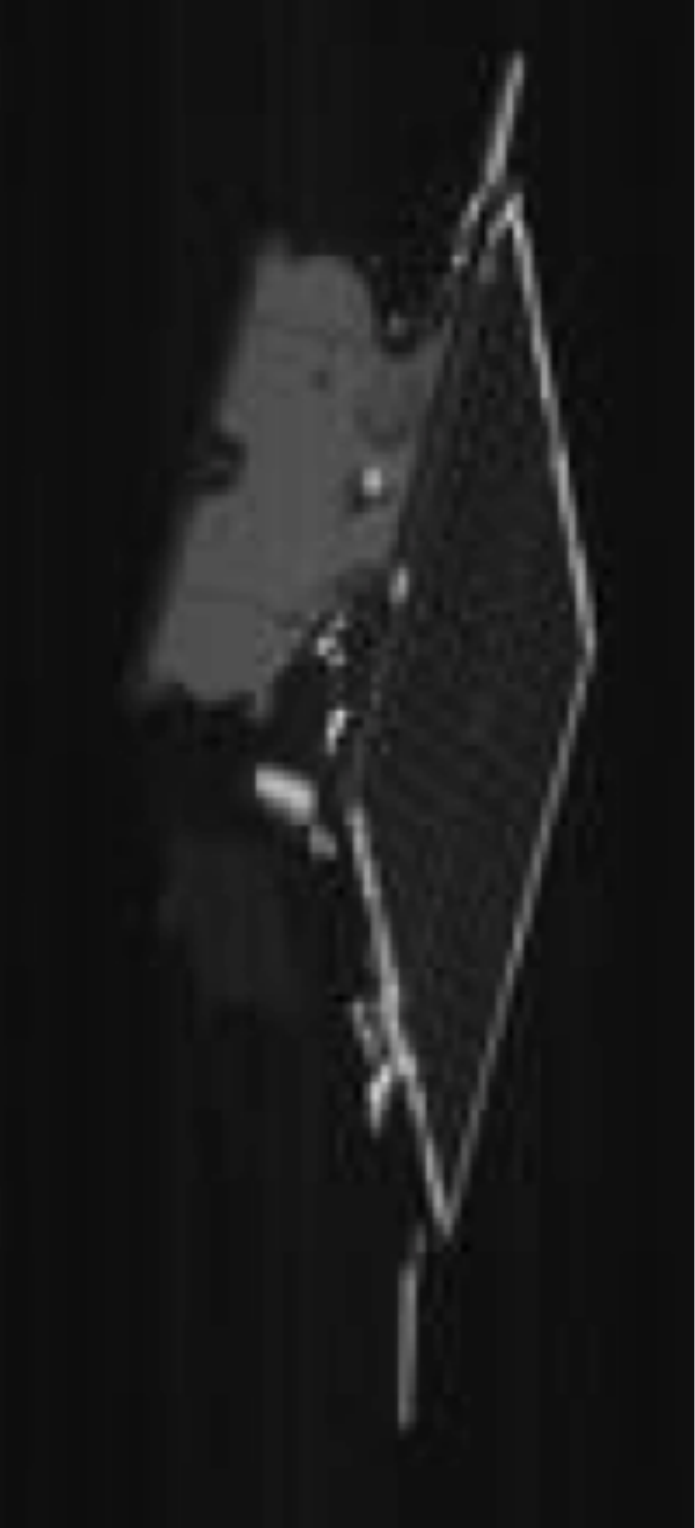}}
	\hfil
	\subfloat[Beierle]{\includegraphics[width=0.16\textwidth]{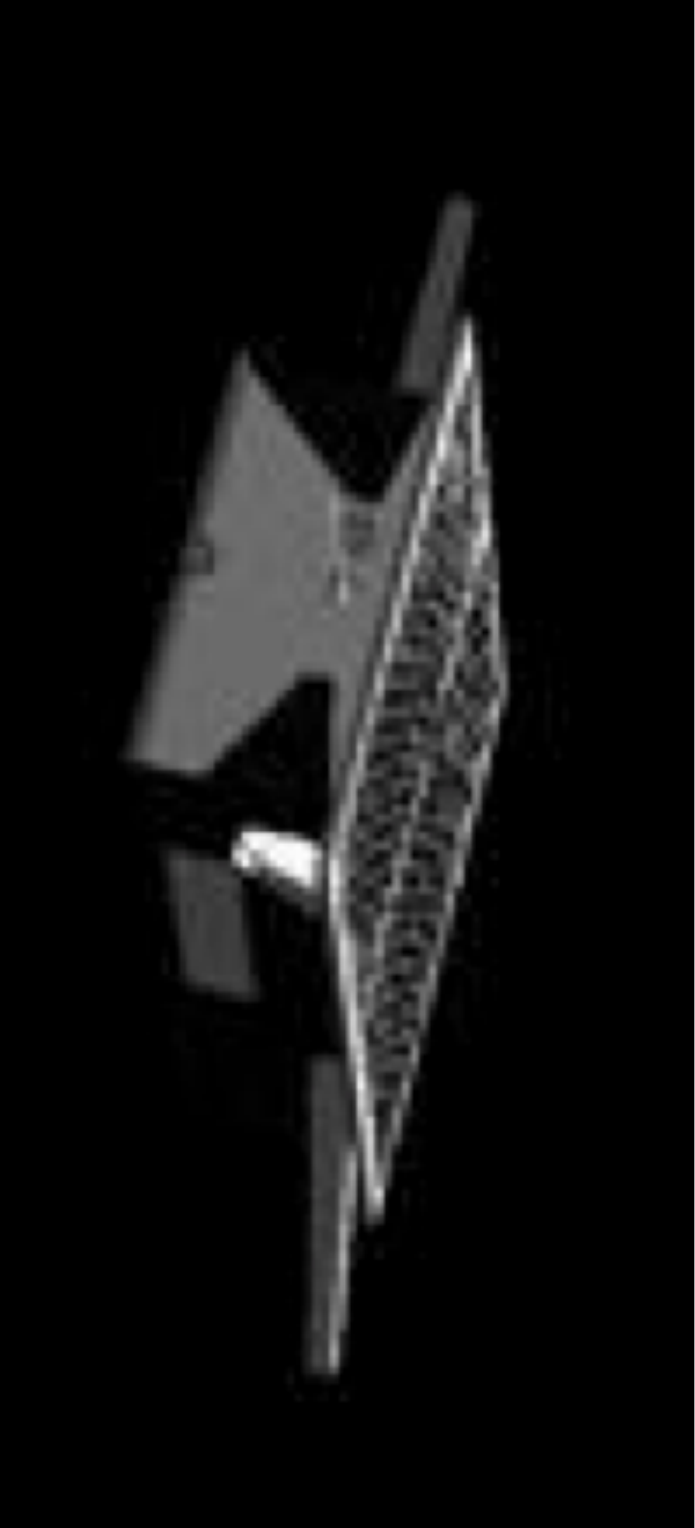}} 
	\hfil 
	\subfloat[SPEED]{\includegraphics[width=0.16\textwidth]{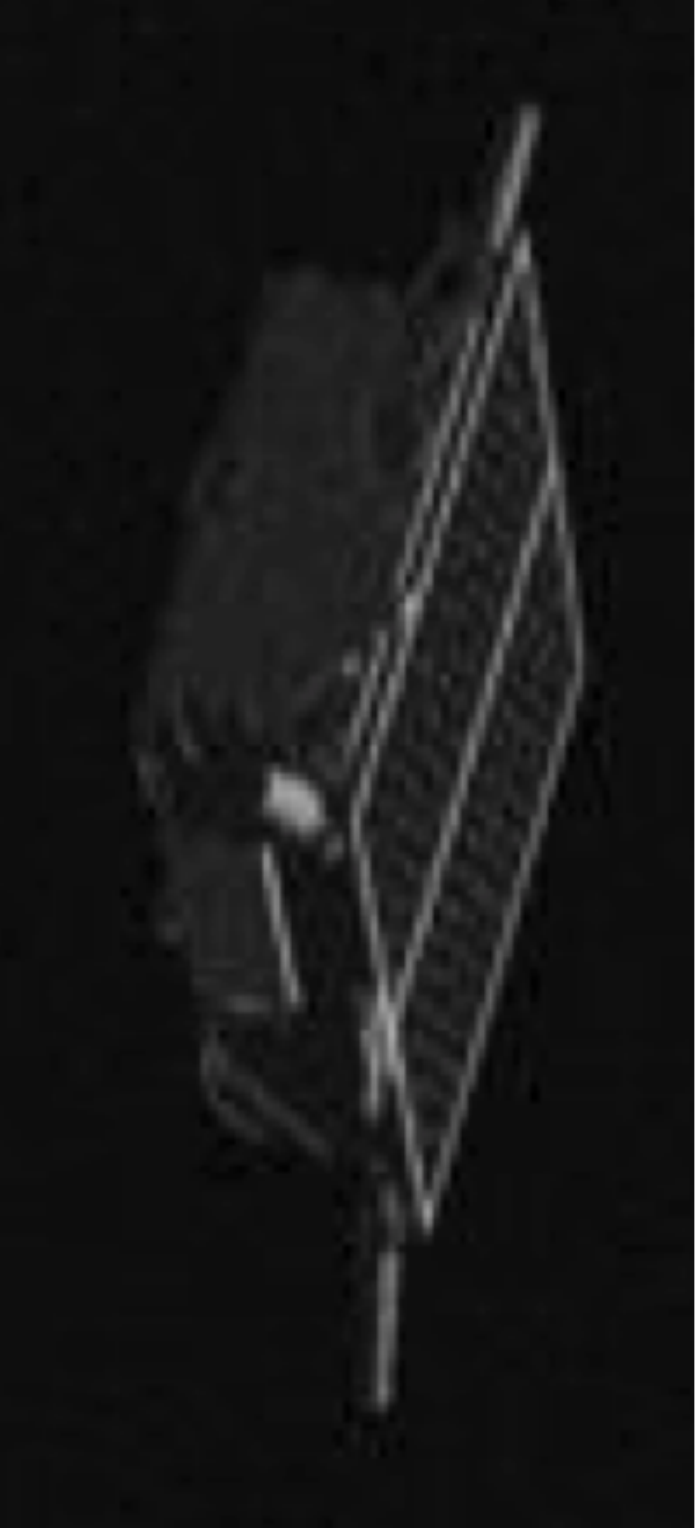}}
	\hfil 
	\subfloat[Histogram of pixel intensities]{\includegraphics[width=0.50\textwidth]{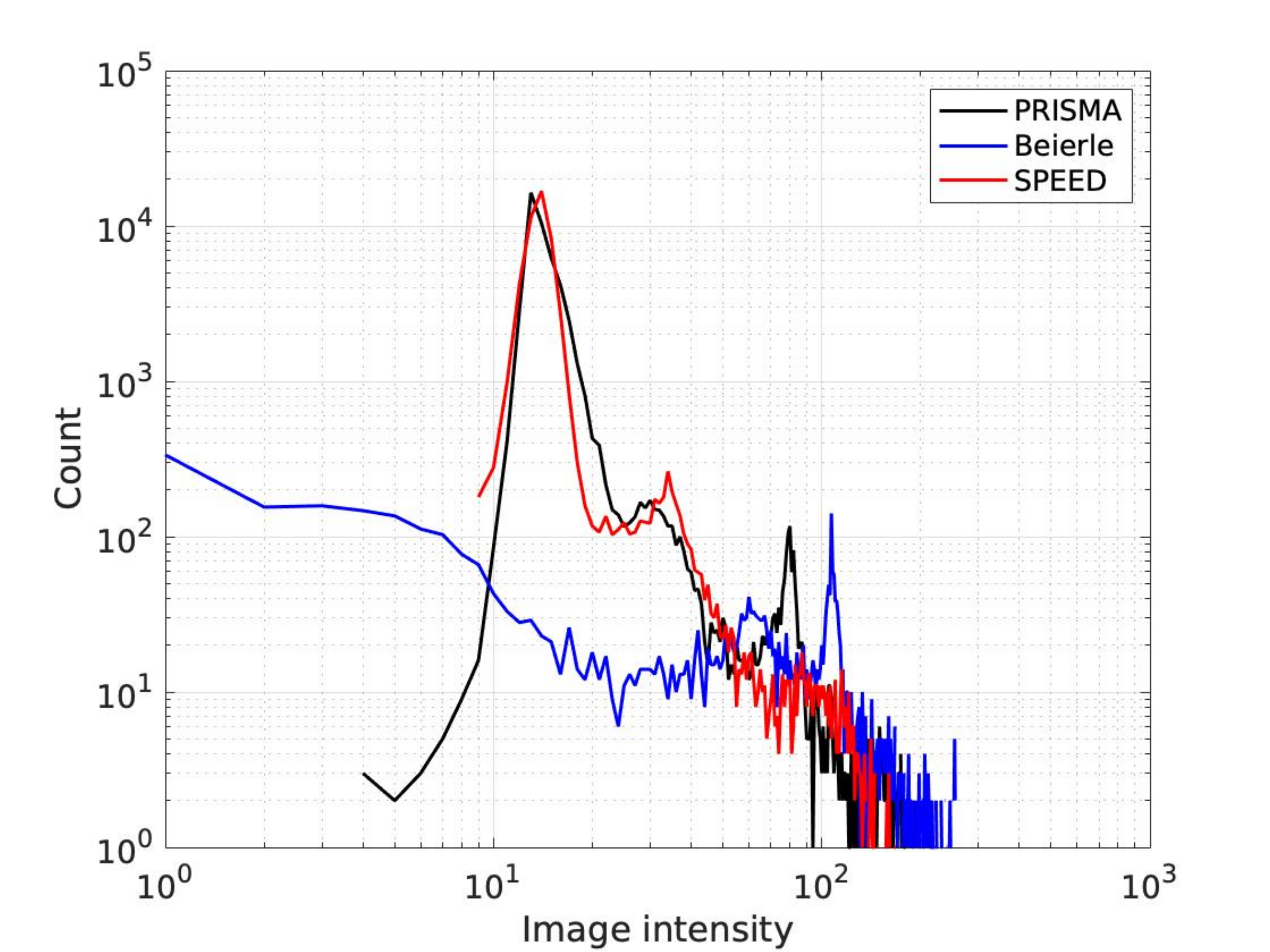}}
	\caption{Cropped versions of (a) the flight imagery captured during the PRISMA mission \cite{PRISMA_chapter}, (b) synthetic imagery in Beierle and D'Amico \cite{Beierle2019}, (c) SPEED synthetic imagery, and (d) histogram comparison of image pixel intensities of the three images. They are cropped from the downscaled 224 $\times$ 224 pixel images.}
	\label{fig:synthetic_comparison}
\end{figure*}

\begin{figure*}[!t]
	\centering
	\includegraphics[width=1\textwidth]{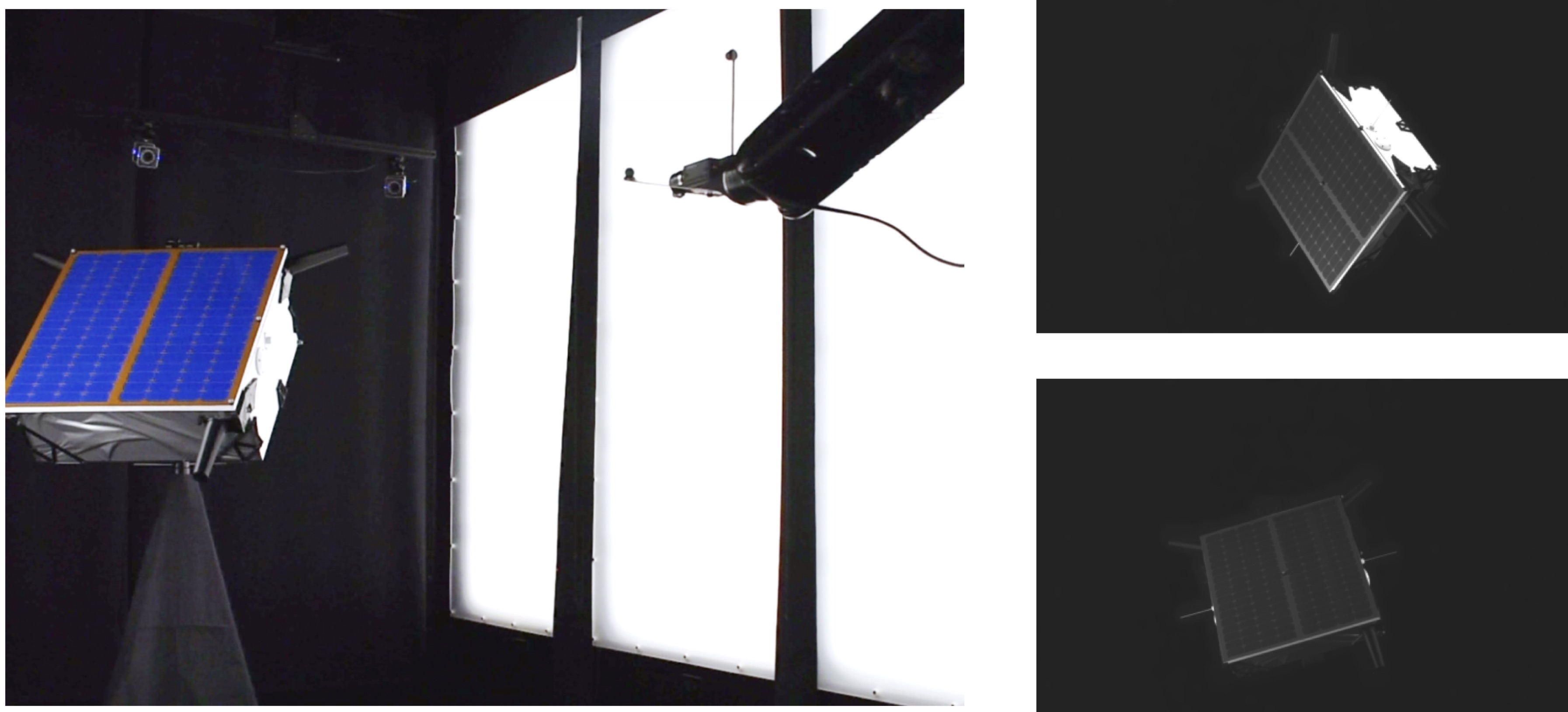}
	\caption{\textit{Left}: TRON facility at SLAB. \textit{Right}: Two examples of real training images from SPEED.}
	\label{fig:TRON2}
\end{figure*}

\begin{figure*}[!t]
	\centering
	\includegraphics[width=1\textwidth]{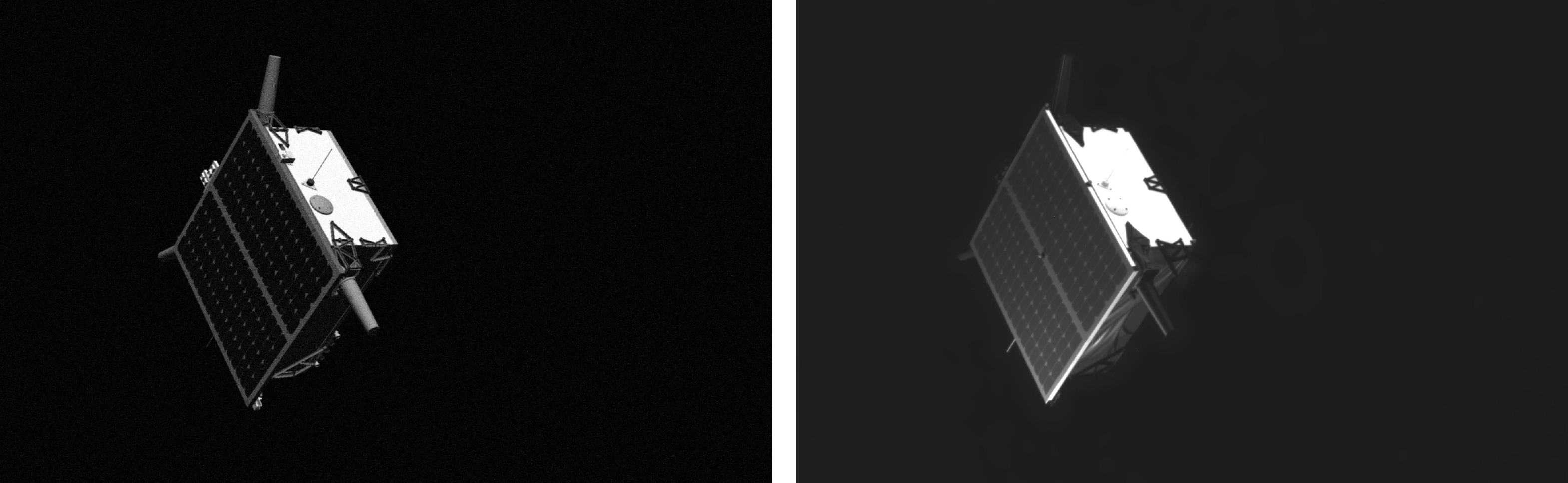}
	\caption{\textit{Left}: SPEED synthetic imagery. \textit{Right}: SPEED real imagery.}
	\label{fig:synth_real_comp}
\end{figure*}

\subsection{Creation of the synthetic dataset}
The synthetic images of the Tango spacecraft are created using the camera emulator software of the Optical Stimulator \cite{Sharma2018_CNN, Beierle2019}. The software uses the OpenGL-based image rendering pipeline to generate photo-realistic images of the Tango spacecraft with desired ground-truth poses (examples are shown on Fig.~\ref{fig:synthetic_examples}). Random Earth images captured by the Himawari-8 geostationary weather satellite\footnote{https://himawari8.nict.go.jp/} are inserted to the background of half of the synthetic images. For these images, the illumination conditions are created to best match those of the background Earth images. Finally, all images are processed with Gaussian blurring ($\sigma = 1$) and zero-mean, Gaussian white noise ($\sigma^2 = 0.0022$) using the MATLAB's \texttt{imgaussfilt} and \texttt{imnoise} commands, respectively.

From Fig.~\ref{fig:synthetic_comparison}, it is clear that the synthetic imagery of SPEED can closely emulate the illumination conditions captured from the actual flight imagery, indicated by the overlapping histogram curves of the image pixel intensities of both imageries. This demonstrates significant improvement of SPEED's image rendering pipeline over the previous work by Beierle and D'Amico \cite{Beierle2019} and its capability of generating photorealistic images of any desired spacecraft with specified pose labels. 

\subsection{Collecting real images with TRON}
The real images of the Tango spacecraft are captured using the TRON facility of SLAB \cite{sharmaThesis2019, Beierle2019} as shown in Fig.~\ref{fig:TRON2}. At the time of image generation, the facility consisted of a 1:1 mockup model of the Tango spacecraft and a ceiling-mounted seven degrees-of-freedom robotic arm, which holds the camera at its end-effector. The facility also includes custom Light-Emitting Diode (LED) wall panels which can simulate the diffused illumination conditions due to Earth albedo and a xenon short-arc lamp to simulate collimated sunlight in various orbit regimes. The ground-truth pose labels for the real images are acquired using ten Vicon cameras \cite{vicon_vero} that track infrared (IR) markers on the satellite mockup and the test camera. Careful calibration processes outlined in \cite{sharmaThesis2019} are performed to remove any biases in the estimated target and camera reference frames. Overall, the independent pose measurement of the calibrated Vicon system provides the pose labels with degree-level and centimeter-level accuracy \cite{sharmaThesis2019}. Current work is undergoing to improve the accuracy of the ground-truth pose by one order of magnitude by fusing Vicon cameras and robot measurements concurrently.

Fig.~\ref{fig:synth_real_comp} provides a qualitative comparison of synthetic and real images of SPEED. Note that while both images share identical ground-truth poses and general direction of Earth albedo, one can readily observe a number of discrepancies in the image properties, such as the spacecraft's texture, illumination and eclipse of certain spacecraft features.

\begin{table}[!t]
	\renewcommand{\arraystretch}{1.2}
	\caption{Number of Images in Different Partitions of the Dataset}
	\label{table:sets}
	\centering
	\begin{tabularx}{0.35\textwidth}{@{}Xbb@{}}
		\toprule
		& Synthetic & Real \\ 
		\midrule
		Training set & 12000  & 5 \\ 
		Test set     & 2998 & 300 \\ 
		\bottomrule
	\end{tabularx}
\end{table}

\subsection{Basic Dataset Properties}
The released dataset contains almost $15000$ synthetic and $300$ real images and is partitioned into the training and test sets according to Table \ref{table:sets}. Note that while synthetic images are partitioned into 8:2 ratio, only five real images are provided with labels for training. It represents a common situation in spaceborne applications in which the images of an orbiting satellite are scarce and difficult to obtain. All images are grayscale with high resolution ($1920 \times 1200$ pixels).

\begin{figure}[!t]
    \centering
    \includegraphics[width=0.6\textwidth]{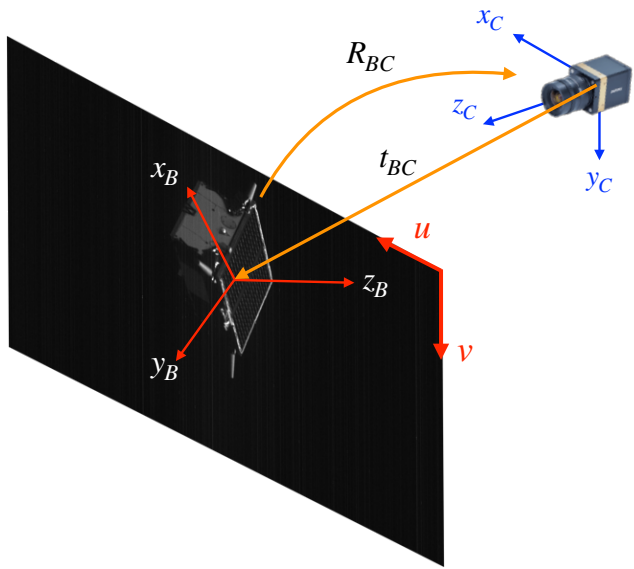}
    \caption{Definition of spacecraft body reference frame ($\mathcal{B}$), camera reference frame ($\mathcal{C}$), relative position ($\bm{t}_\mathcal{BC}$), and relative orientation ($\bm{R}_\mathcal{BC}$).}
    \label{fig:Reference Frames}
\end{figure}

\begin{figure*}[!t]
	\centering
	\includegraphics[width=1\textwidth]{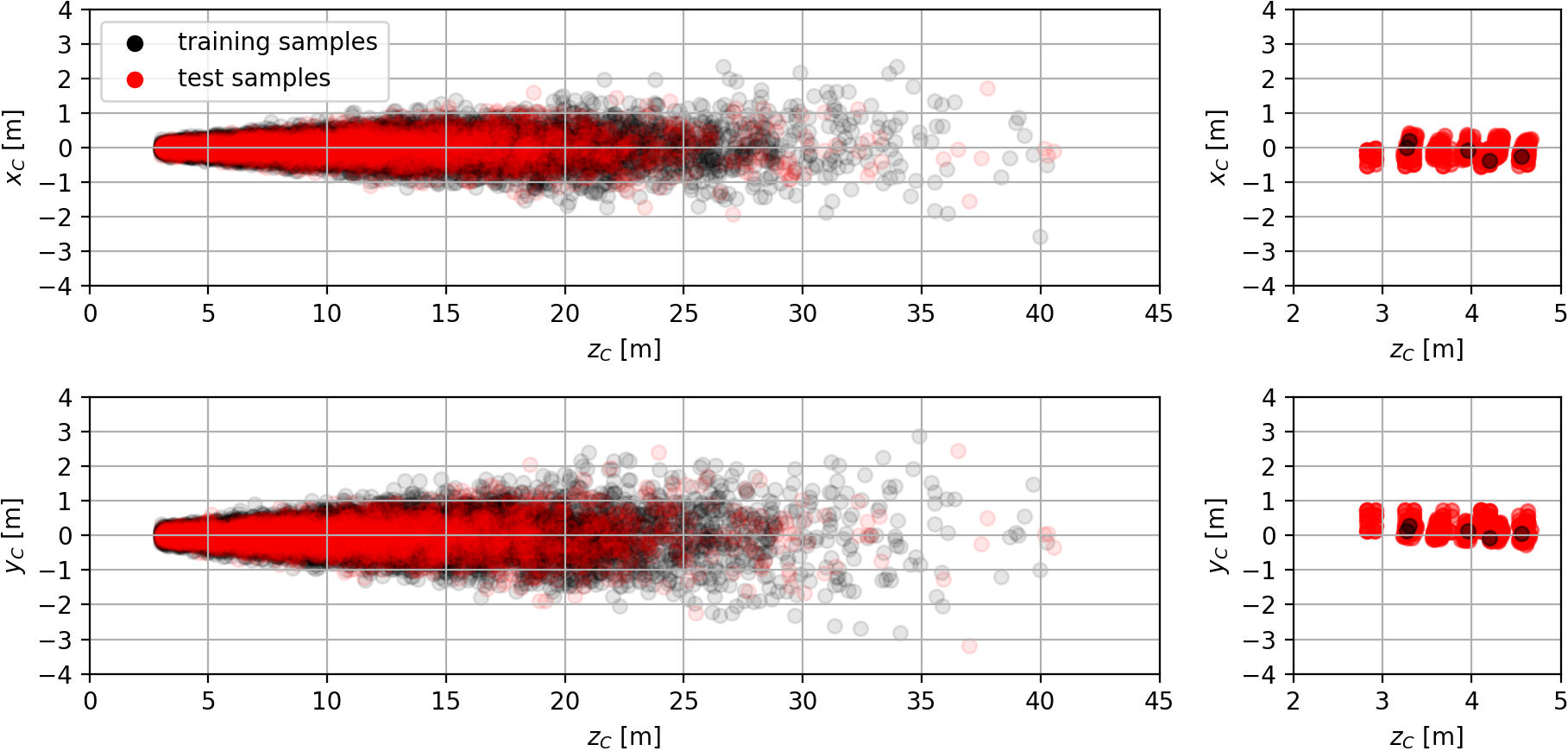}
	\caption{Position distributions of the pose labels across the dataset in the camera frame ($\mathcal{C}$), for synthetic (\textit{left}) and real (\textit{right}) samples.}
	\label{fig:position_distr}
\end{figure*}

Fig.~\ref{fig:Reference Frames} graphically describes the spacecraft body and camera reference frames to visualize the position and orientation distributions of the dataset. Specifically, $z_\mathcal{C}$ is aligned with the camera boresight in the camera reference frame, while $z_\mathcal{B}$ is perpendicular to the solar panel in the Tango's body reference frame. ($x_\mathcal{C}$, $y_\mathcal{C}$) and ($x_\mathcal{B}$, $y_\mathcal{B}$) then form a plane perpendicular to $z_\mathcal{C}$ and $z_\mathcal{B}$, respectively, as shown in Fig.~\ref{fig:Reference Frames}.

\begin{figure*}[!p]
	\centering
	\includegraphics[width=1\textwidth]{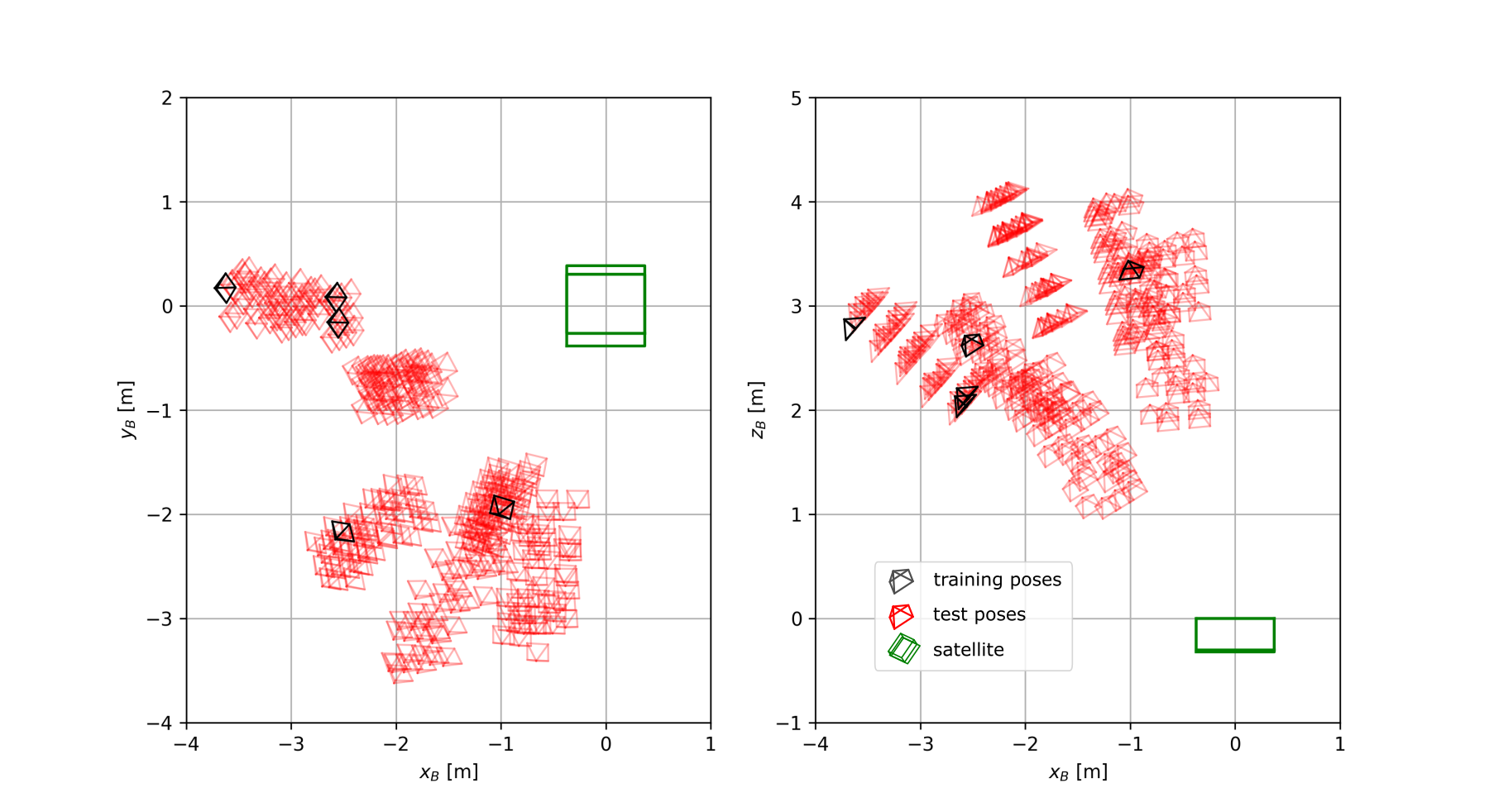}
	\caption{Camera poses for real images in the Tango's body frame ($\mathcal{B}$) from two views. The simplified wireframe model of the satellite is plotted in green, camera poses are plotted in red and black for test and training samples, respectively.}
	\label{fig:pose_distr_real}
\end{figure*}

\begin{figure*}[!p]
	\centering
	\includegraphics[width=1\textwidth]{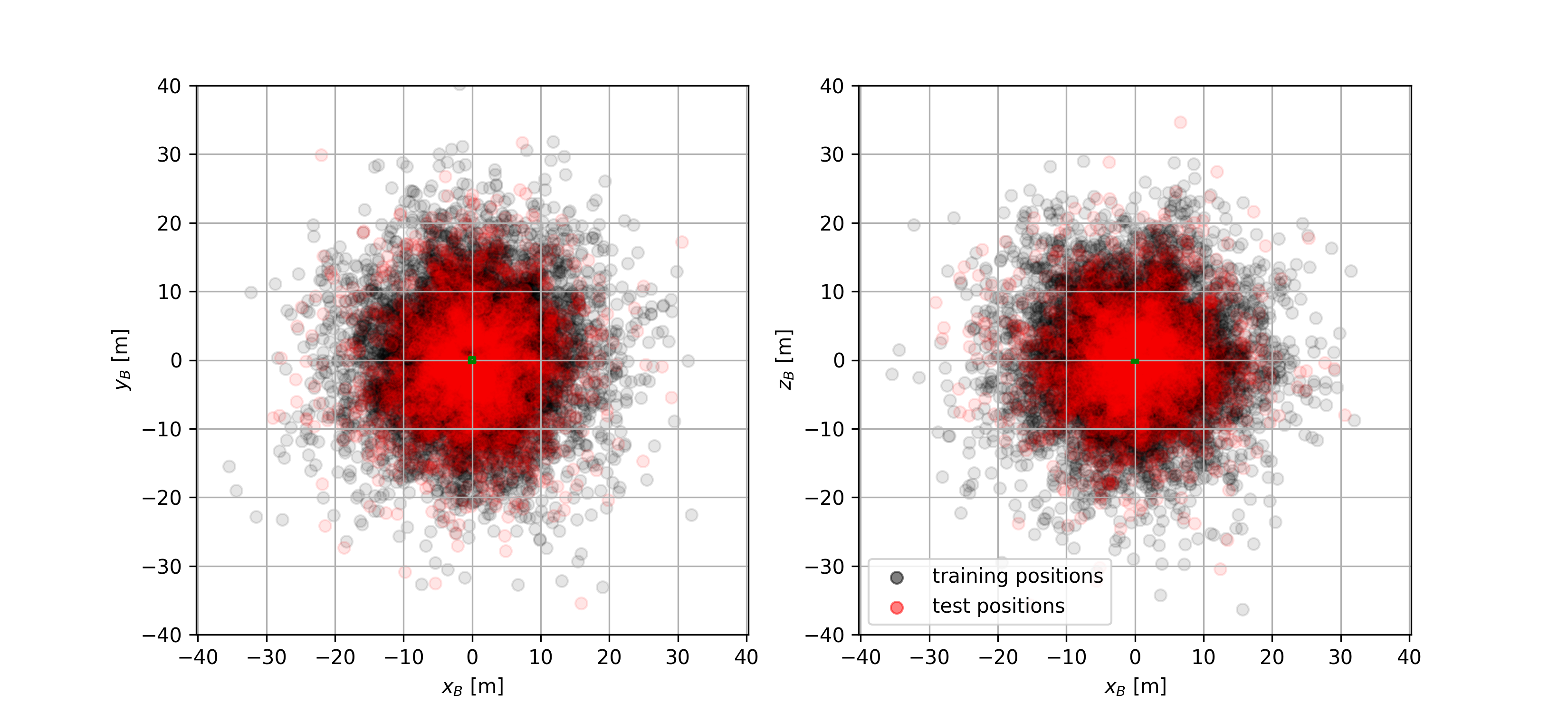}
	\caption{Distribution of the camera's relative positions of the synthetic images in the Tango's body frame ($\mathcal{B})$ from two views. The satellite is in the origin, training and test camera poses are plotted in red and black for test and training samples respectively.}
	\label{fig:pose_distr_syn}
\end{figure*}

Fig.~\ref{fig:position_distr} shows the range of relative position distributions in the dataset in the camera frame. The distance of the satellite in the synthetic images is between 3 and 40.5 meters. Due to physical limitations of the TRON facility in combination with the size of the satellite mockup, the distance distribution of real images is much more constrained, ranging from 2.8 to 4.7 meters.

Fig.~\ref{fig:pose_distr_real} visualizes the relative orientation and position distributions for real images in the satellite body frame. For synthetic images, Fig.~\ref{fig:pose_distr_syn} visualizes the relative position distribution in the satellite body frame. It especially visualizes the fact that for synthetic images, the relative orientations are well distributed across the 3D space. However, in case of real images, the diversity of orientations and distances is restricted due to physical limitations.

\section{Competition Design}
\label{sec:competition_design}
In an open scientific competition such as SPEC and other Kelvins competitions, scientific problems are turned into well-formulated mathematical problems that are solved by engaging the broader scientific community and citizen scientists. Therefore, there are two key factors that are considered in setting up the competition:
\begin{itemize}[
    \setlength{\IEEElabelindent}{\dimexpr-\labelwidth-\labelsep}
    \setlength{\itemindent}{\dimexpr\labelwidth+\labelsep}
    \setlength{\listparindent}{\parindent}
    ]
    \item \textbf{community engagement}: The participants and the effort they put into solving the problems are our main resource. Therefore, a broad audience has to be reached to attract many individuals and teams. Then, the barrier to entry into the competition has to be as low as possible. Finally, engagement of the participants has to be maintained. This last point involves making sure that the problem can be solved based on the released dataset (e.g., images contain sufficient information for pose estimation, samples are well distributed in the pose space, etc.), that solutions are quickly evaluated and added to a live leader board, and in general that the competition is fair (e.g., by keeping the test set private).
    
    \item \textbf{competition metric}: The creation of the competition metric is the process in which the scientific problem of interest is turned into an optimization problem. Care should be taken in designing the competition metric, as it has to directly reflect the important aspects of the problem. Otherwise a solution to the optimization problem might not be relevant to the original scientific problem. In case the metric can be cheated, participants may focus on specific solutions that might lead to good scores but are of less practical value.
    
\end{itemize}

SPEC particularly aimed to focus community efforts on the problem of estimating pose of uncooperative satellites. The following sections describe the competition setup and the baseline solutions provided to the participants and introduce the competition metric.

\subsection{Competition setup - the Kelvins competition platform}

Kelvins, the platform which hosts SPEC and many other satellite-related challenges, was designed to provide a seamless experience for the participants. It features a live leaderboard that is a key for maintaining community engagement over longer intervals. Teams have direct information about how their latest submission compares to their peers, the limits are constantly pushed further, and the competitive aspect brings more motivation for teams to put in effort. Another important feature is the automated evaluation of submissions. This allows for keeping the test set private, which helps ensuring a fair competition. During the competition only $20\%$ of the test set was used for evaluation and placement in the leaderboard in order to prevent the participants from overfitting on the entire test set.

\subsection{Competition Metric}
The competition metric has to faithfully reflect the underlying scientific problem in order to ensure that the high-scoring solutions are meaningful also outside the context of the competition. While it is not uncommon to have separate orientation and position metrics \cite{Kendall2015_PoseNet}, a single scalar score was used instead to rank the submissions on the leaderboard.

To evaluate the submitted pose solutions, separate position ($e_\textrm{t}$) and orientation ($e_\textrm{q}$) errors are computed. Fig.~\ref{fig:Reference Frames} graphically describes the relevant reference frames to compute the errors. The position error, $e_\textrm{t}$, is defined as
\begin{equation} \label{eq:pos_error}
    e_\textrm{t} = \left| \bm{t_\mathcal{BC}} - \bm{\hat{t}_\mathcal{BC}} \right|_{2},
\end{equation}
the magnitude (2-norm) of difference between the ground-truth ($\bm{t_\mathcal{BC}}$) and estimated ($\bm{\hat{t}_\mathcal{BC}}$) position vectors from the origin of the camera reference frame $\mathcal{C}$ to that of the target body frame $\mathcal{B}$. The normalized position error, $\bar{e}_\textrm{t}$ is also defined as 
\begin{equation} \label{eq:pos_error_norm}
    \bar{e}_\textrm{t} = \frac{e_\textrm{t}}{\left| \bm{t_\mathcal{BC}} \right|_{2}},
\end{equation}
which penalizes the position errors more heavily when the target satellite is closer.

The orientation error $e_\textrm{q}$ is calculated as the angular distance between the predicted, $\bm{\hat{q}}$ = $\bm{q}(\bm{\hat{R}_\mathcal{BC}})$, and true, $\bm{q}$ = $\bm{q}(\bm{R_\mathcal{BC}})$, unit quaternions, i.e., the magnitude of the rotation that aligns the target body frame $\mathcal{B}$ with the camera reference frame $\mathcal{C}$,
\begin{equation} \label{eq:orientation_error}
    e_\textrm{q} = 2  \cdot  \arccos\left( \left| \left< \bm{\hat{q}}, \bm{q}\right> \right| \right).
\end{equation}
where $\left< \bm{\hat{q}}, \bm{q} \right>$ denotes the inner product of two unit quaternions.

The pose error $e_\textrm{pose}$ for a single image is the sum (1-norm) of the orientation and the normalized position error,
\begin{equation} \label{eqn:pose_error_single}
    e_\textrm{pose} = e_\textrm{q} + \bar{e}_\textrm{t},
\end{equation}

\noindent Finally, the total error $E$ is the average of the pose errors over all $N$ images of the test set,
\begin{equation}  \label{eq:total_error}
    E = \frac 1N \sum^{N}_{i=1}e_\textrm{pose}^{(i)}.
\end{equation}

A main concern during the creation of the competition metric was to balance its sensitivity to position and orientation errors and avoid situations where one factor dominates the metric while neglecting the other. Note that since the position error is dependent on the target distance, the balance between the two contributions also depends on the particular distance distribution of the test set.

In order to check the balance of the sensitivities, the total error $E$ was calculated over the test set for two cases: introducing $1^\circ$ of orientation error in the first case, and adding $0.1$ m translation error in the second case. It was shown that $0.1$ m translation error, on average, is equivalent to $0.7094^\circ$ error for the particular distribution of poses in the test in the first case. Likewise, $1^\circ$ orientation error was shown to be equivalent to $0.141$ m translation error in the second case. Such behavior is expected due to the underlying perspective equations which drive image formation. This suggested the contributions of each error type are reasonably balanced, thus the total score combines both errors without the introduction of additional scaling factors. 

Two alternative metrics were also considered. The reprojection error is the average distance between projected keypoints measured in 2D on the image plane \cite{Brachmann2016UncertaintyDriven6P}. The average distance error is the 3D distance between the ground truth and predicted keypoints (usually referred to as ADD metric \cite{Hinterstoisser2013_LINEMOD}). Both have the disadvantage that the orientation and position sensitivity is dependent on the choice of keypoints, since the slope of orientation error is proportional to the distance of the keypoints from the origin of the target's body frame. Furthermore, the reprojection error is numerically unstable in the case when predicted keypoints lie very close to the image plane. 

\subsection{Baseline solutions}

Two different example solutions are provided to the participants in Python using two popular deep learning frameworks, Keras and PyTorch\footnote{https://gitlab.com/EuropeanSpaceAgency/speed-utils}. The main reason for providing these baseline solutions is to lower the barriers of entering the competition. While the performance of these baselines is intentionally rather weak, it still allows competitors to submit their first result within an hour. Along with the example solutions, the competition platform provides useful tools that facilitate working with the dataset, such as functions to visualize samples and corresponding pose labels, or data loaders for the two deep learning frameworks.

The baseline solutions rely on pre-trained ResNet models \cite{He2015_ResNet} where the last layer is replaced with a layer containing seven linear outputs for the pose variables. The models are fed with downscaled $224 \times 224$ pixel images and trained with simple Mean-Squared Error (MSE) loss for $20$ epochs. These baselines leave quite some room for improvements. For instance, the outputs are not normalized, or the predicted distance along the camera boresight is typically one order of magnitude larger than all the other output variables. Using the MSE loss, errors in this direction dominate the loss. Furthermore, MSE loss does not account for the periodicity of orientation.

Keeping the baseline solutions intentionally simple and weak helped to engage the participants in the competition. These baselines allow for incremental improvements, such as replacing the loss function or training on larger input images. Additionally, a stronger third baseline solution, also based on CNN, was developed during the competition by SLAB and is used for comparison purposes. 

\section{Competition Results}
\label{sec:competition_results}

During the competition, 48 teams participated and submitted results. 20 teams filled a post-competition questionnaire and provided detailed descriptions about their approaches. This section analyzes and compares their submissions, provides a brief description of the top 4 approaches, evaluates the performance of the different approaches, and identifies difficult samples to show what are the current limits of this technology. 

\subsection{Final results}

\begin{figure*}[!t]
	\centering
	\includegraphics[width=0.8\textwidth]{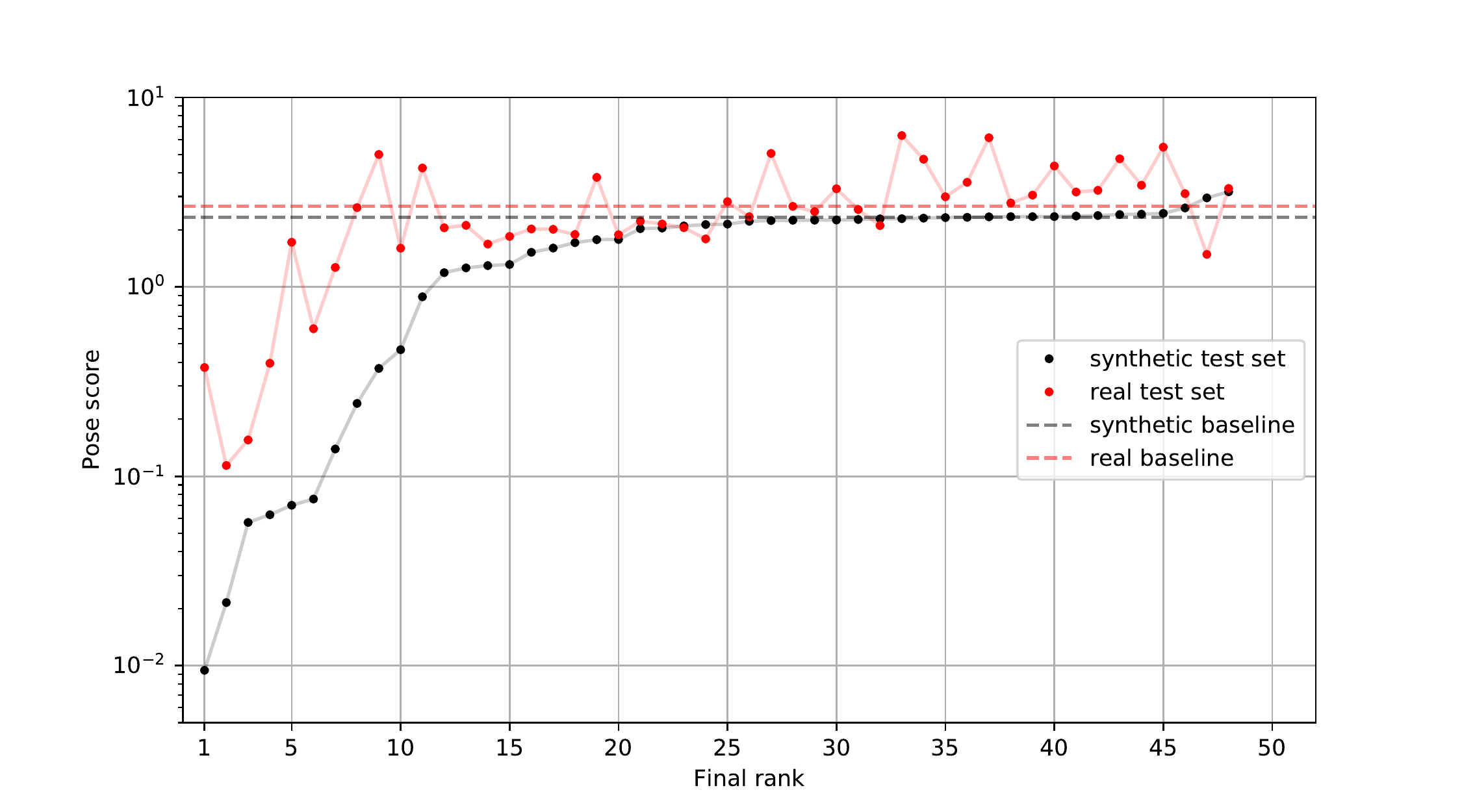}
	\caption{Final results on the synthetic and real test sets.}
	\label{fig:final_results}
\end{figure*}

\begin{table*}[!t]
	\centering
	\begin{threeparttable}
		\renewcommand{\arraystretch}{1.2}
		\caption{Detailed Results of the Top Ten Submissions Compared to the SLAB's Baseline Performance}
		\label{table:scores}
		\begin{tabularx}{0.8\textwidth}{@{}Bmmbbs@{}}
			\toprule
			Team & $E_\textrm{syn}$  & $E_\textrm{real}$ & $e_\textrm{q}$ $[^\circ]$  & $e_\textrm{t}$ {[}m{]}  & PnP \\
			\midrule
			1. UniAdelaide \cite{Chen2019SPEC} & \textbf{0.0094} & 0.3752 & $\bm{0.41 \ \pm \ 1.50}$ & $\bm{0.032 \ \pm \ 0.095}$ & Yes \\
			2. EPFL\_cvlab & 0.0215 & \textbf{0.1140} & $0.91 \ \pm \ 1.29$ & $0.073 \ \pm \ 0.587$ & Yes \\
			3. pedro\_fairspace \cite{Proenca2019SPEC} & 0.0571 & 0.1555 & $2.49 \ \pm \ 3.02$ & $0.145 \ \pm \ 0.239$ & No \\  \hline
			SLAB Baseline \cite{Park2019_TowardsRL} & 0.0626 & 0.3951 & $2.62 \ \pm \ 2.90$ & $0.209 \ \pm \ 1.133$ & Yes \\  \hline
			4. Team\_Platypus & 0.0703 & 1.7201 & $3.11 \ \pm \ 4.31$ & $0.221 \ \pm \ 0.530$ & No \\
			5. motokimura1 & 0.0758 & 0.6011 & $3.28 \ \pm \ 3.56$ & $0.259 \ \pm \ 0.598$ & No \\
			6. Magpies & 0.1393 & 1.2659 & $6.25 \ \pm \ 13.21$ & $0.314 \ \pm \ 0.568$ & No \\
			7. GabrielA & 0.2423 & 2.6209 & $12.03 \ \pm \ 12.87$ & $0.318 \ \pm \ 0.323$ & No \\
			8. stainsby & 0.3711 & 5.0004 & $17.75 \ \pm \ 22.01$ & $0.714 \ \pm \ 1.012$ & No \\
			9. VSI\_Feeney & 0.4658 & 1.5993 & $23.42 \ \pm \ 33.57$ & $0.734 \ \pm \ 1.273$ & No \\
			10. jblumenkamp & 0.8859 & 4.2418 & $35.92 \ \pm \ 49.72$ & $2.656 \ \pm \ 2.149$ & Yes \\
			\bottomrule
		\end{tabularx}
		\begin{tablenotes}
			Best results for each metric are highlighted with bold fonts. The mean and the standard deviation of the orientation errors ($e_\textrm{q}$) as in (\ref{eq:orientation_error}) and position errors ($e_\textrm{t}$) as in (\ref{eq:pos_error}) are measured on the synthetic test set.
		\end{tablenotes}
	\end{threeparttable}
\end{table*}

Fig.~\ref{fig:final_results} illustrates the final scores. The first 20 teams significantly outperformed the initial baseline with the top teams getting a two orders of magnitude improvement over the baseline solutions.\footnote{Final leaderboard: https://kelvins.esa.int/satellite-pose-estimation-challenge/results/}

While the primary competition ranking criteria was the score on the synthetic test set, submissions were also evaluated on the real test set. Results on real images are weaker compared to those on synthetic images for most teams, except for three of the solutions. Machine learning models are generally expected to perform worse when evaluated on data with a statistical distribution that significantly differs from their training set. It is possible that the reason those three teams achieved better results on real imagery is related to its limited pose distribution.

The results of the top ten teams are collected in Table \ref{table:scores} and compared to the baseline network developed by SLAB \cite{Park2019_TowardsRL} during the course of the competition. While team \texttt{UniAdelaide} \cite{Chen2019SPEC} won the competition by achieving the highest score on the synthetic test set, \texttt{EPFL\_cvlab}\footnote{https://www.epfl.ch/labs/cvlab/} achieved the highest accuracy on real images. \texttt{pedro\_fairspace} \cite{Proenca2019SPEC} submitted the best submission that did not rely on PnP solvers, finishing on the third place. These top three solutions were the only submissions to outperform the SLAB baseline. Before the competition, the best published result on SPEED was Spacacraft Pose Network (SPN) by Sharma and D'Amico \cite{Sharma2019, sharmaThesis2019}. SPN was also the first published result on SPEED benchmark prior to its public release, and its reported performances in terms of the mean orientation and position error are $e_\textrm{q}=8.43^\circ$ and $e_\textrm{t}=0.783$ m.

\subsection{Approaches of the top 4 competitors}

Here, the technical details of the top 4 competitors (i.e., top 3 performers and the SLAB baseline) are briefly covered for completeness. For detailed descriptions of each method, the readers are encouraged to refer to the relevant materials.
\begin{enumerate}
    \item \texttt{UniAdelaide} \cite{Chen2019SPEC}: The team \texttt{UniAdelaide} first recovers the 3D coordinates of arbitrarily chosen 11 landmark points, or keypoints, on the Tango satellite via multi-view triangulation. They also regress the 2D bounding box around the satellite using an object detection CNN, which is used to crop the satellite from an original image. The bounding box labels for the training images are obtained by projecting the recovered landmarks onto 2D image plane using the provided ground-truth pose labels. Then, the team trains a landmark regression CNN on cropped images to obtain the 2D locations of all 11 landmarks. The team uses HRNet \cite{Sun2019HRNet} to regress the heatmaps associated with each landmark instead of directly regressing the 2D coordinates. Finally, given predicted 2D-3D correspondences of the landmarks, they perform a robust nonlinear optimization to compute the pose estimates.
    
    \item \texttt{EPFL\_cvlab}\footnote{https://indico.esa.int/event/319/attachments/3561/4754/pose\_gerard\_segmentation.pdf}: Similar to the approach of the team \texttt{UniAdelaide}, the team \texttt{EPFL\_cvlab} regresses 8 landmarks corresponding to 8 corners of the satellite's cubic body. Their approach is based on segmentation-driven CNN \cite{Hu2019SegmentationDrive}, which divides the image into $S \times S$ grids and has each grid predict the presence of the object (segmentation) and 2D locations of the 8 keypoints along with their confidence values. Then, out of all keypoint candidates, $n$ most confident keypoints are used to compute the pose estimates using a RANSAC-based PnP solver \cite{Fischler1987RANSAC} . 
    
    \item \texttt{pedro\_fairspace} \cite{Proenca2019SPEC}: The team adopts the ResNet-based architecture \cite{He2015_ResNet} to directly regress the 3D position and unit quaternion vectors. Such mechanism allows for directly optimizing the competition pose error as defined in Eq.~\ref{eqn:pose_error_single}. Instead of a norm-based loss of unit quaternions, the team proposes to formulate the orientation regression as soft classification based on a Gaussian mixture model to handle the attitude ambiguity.
    
    \item \texttt{SLAB Baseline} \cite{Park2019_TowardsRL}: The team \texttt{SLAB Baseline} exploits a pose estimation architecture similar to that of the team \texttt{UniAdelaide}, incorporating the recovery of 11 keypoints and a separate object detection CNN to crop the most relevant areas of the images. They use YOLO-based CNN architectures \cite{redmonFarhadi2017YOLOv2, redmonFarhadi2018YOLOv3} for both object detection and keypoint regression tasks. However, the architectures exploit depth-wise separable convolution operations \cite{Sandler_MobileNet_v2} to significantly reduce the number of parameters associated with each network. They use EPnP \cite{Lepetit2008} to compute the pose estimates.
\end{enumerate}

\subsection{Survey on methods}
\label{sec:survey}

Shortly after the competition, all participants were asked to answer a short surveying questionnaire regarding their backgrounds, the approaches they used, and how they dealt with certain aspects of the problem. 20 teams, including the top 13 competitors, answered the survey. Most of the teams (except for three) consisted of a single individual contributor, affiliated with academic institutions ($35\%$) or industry ($30\%$). It is noteworthy that only half of the teams were involved with space related research, and $65\%$ were not working on pose estimation problems at all.

Deep learning approaches dominated the submissions, as all teams used deep learning either in an end-to-end fashion or as an intermediate process in their pipelines. The teams addressed the pose estimation problem as a regression task, except for one team that framed orientation prediction as a soft classification problem. Various architectures were used from well known pre-trained models, such as ResNets \cite{He2015_ResNet}, Inception v3 \cite{Szegedy2015Inceptionv3}, and YOLO \cite{redmonFarhadi2017YOLOv2, redmonFarhadi2018YOLOv3}, to custom models trained from scratch. 18 of the 20 teams made use of the data augmentation techniques to maximize their performance, such as geometric transformations (e.g., rotation around the camera axis, zooming and cropping) and pixel intensity changes (e.g., adding noise, changing brightness).

SPEED consists of high resolution images that are not suitable as direct inputs to a neural network due to memory limitations of GPUs. Therefore, all teams performed downscaling of the given images to a variety of sizes ranging from $224 \times 224$ to $960 \times 640$ pixels. Some teams cropped the input image, either taking a sufficiently large central crop or localizing the satellite first and then cropping the relevant part of the image. Specifically, a number of top-scoring teams used a separate CNN to perform localization before cropping in order to prevent any loss of information due to downscaling. $60\%$ of the teams used ImageNet pre-trained models that expect three channel RGB input images. Since the dataset consists of single channel grayscale images, this provided additional freedom for teams for constructing their input. While most teams simply stacked the same input channel to have RGB input, two teams included masked or filtered versions of the input on the extra channels.

\begin{figure*}[!t]
	\centering
	\includegraphics[width=1\textwidth]{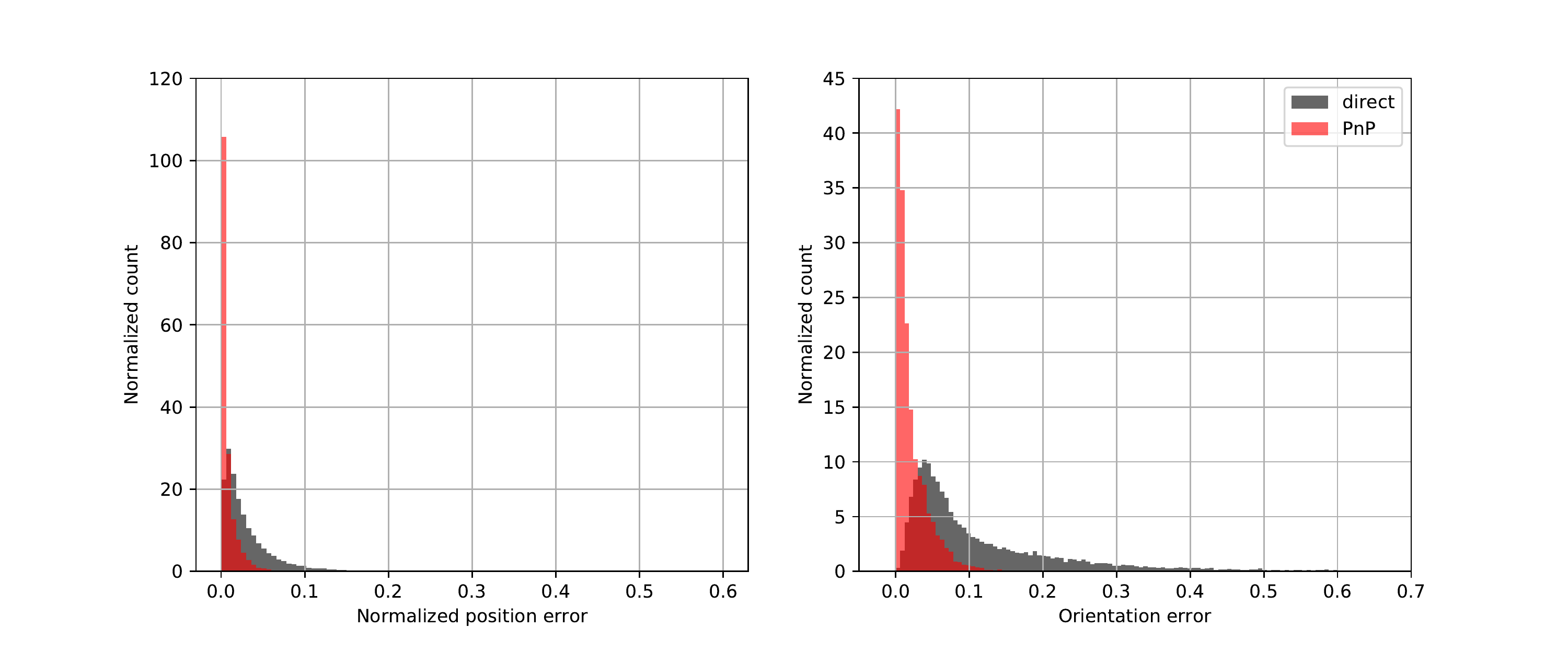}
	\caption{Position (\textit{left}) and orientation (\textit{right}) error distributions for direct and PnP solver based methods.}
	\label{fig:direct_vs_pnp}
\end{figure*}

\begin{figure*}[!t]
	\centering
	\includegraphics[width=1\textwidth]{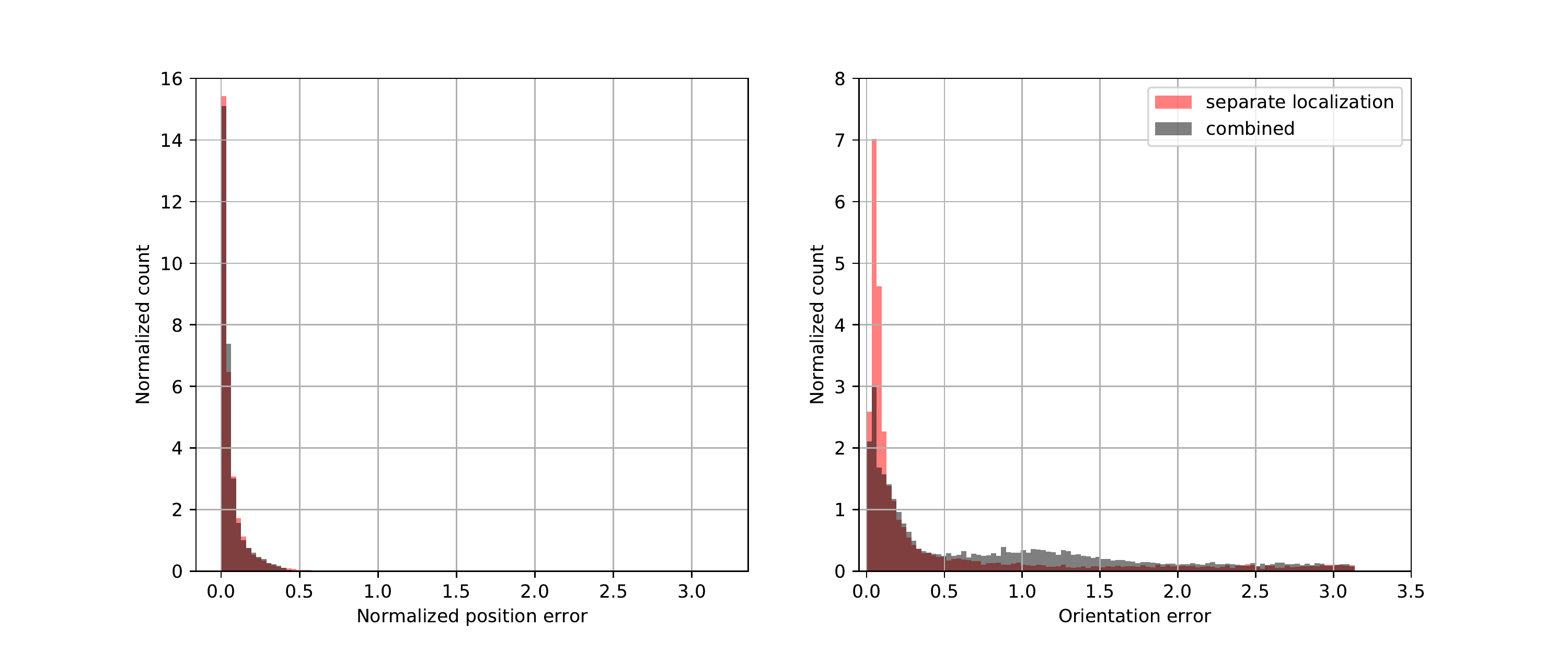}
	\caption{Position (\textit{left}) and orientation (\textit{right}) error distributions highlighting the effect of a localization step prior to orientation estimation.}
	\label{fig:combined_vs_localization}
\end{figure*}

Since the 3D model for the satellite was not released as part of the competition, some teams chose to reconstruct the satellite model in order to use any keypoints-based architecture. Specifically, seven teams reconstructed the 3D coordinate locations of 8 to 11 keypoints using 10 to 20 hand-selected images and the provided pose labels. The keypoints generally correspond to the corners of the satellite body and the tips of the antennae. The method of reconstruction ranged from manually aligning the vertices to triangulation or reprojection-based optimization. The resulting models were used for generating bounding box or segmentation ground truth from the available pose labels, and in some cases directly in the pose estimation process with PnP solvers.

\subsection{Comparing approaches}
\label{sec:approaches}
 
This section provides the analysis of survey results and submissions together to compare design decisions in light of the final results. In particular, it discusses how keypoint matching techniques compare to pure deep learning approaches and what the effect of a separate localization step is in the pose estimation pipeline.
 
\subsubsection{Keypoint matching techniques}

Most teams designed an architecture that predicts the target's pose in an end-to-end fashion. However, four teams designed an architecture that first predicts a set of pre-defined keypoints using a neural network. Then, they use a keypoint matching technique such as a PnP solver to align a known model of the satellite (e.g., reconstructed 3D keypoint coordinates) with the detected keypoints. While the PnP optimization is prone to local minima, it allows for explicitly incorporating the geometric constraints in the pose estimation process.

Fig.~\ref{fig:direct_vs_pnp} illustrates the error distributions for the solutions based on PnP and direct pose estimation separately for position and orientation error. Specifically, the performances of the top 10 teams were analyzed to compare the PnP solutions and strong direct pose estimation submissions. In the submissions, PnP-based solutions significantly outperform direct pose estimation both in terms of position and orientation performance, ranking on the first, second and fourth places. The average orientation errors and their deviations are  $9.76 ^\circ \pm  18.51^\circ$ and $1.31 ^\circ \pm  2.24^\circ$ for direct and PnP methods, respectively, while relative position errors are $0.0328 \pm 0.0430$ m and $0.0083 \pm 0.0269$ m.

\subsubsection{The effect of separate localization}

Another recurring technique across the participants is the use of a separate localization step. In this case, the first step is the detection of the satellite, either by segmenting its contour or identifying a tight-fitting bounding box around it. This step separates the position and orientation estimation tasks, and allows to train separate models. The main advantage is that an intermediate detection result allows for cropping the original high resolution image, to use only the relevant part of the images downstream. The disadvantages of this approach are the added complexity and the need for segmentation/bounding box annotation via a separate model reconstruction step. 

Fig.~\ref{fig:combined_vs_localization} compares the error distributions of the top 8 teams that use direct pose estimation methods (i.e., no PnP solver). Specifically, the half of the selected teams uses an independent localization step in their direct pose estimation approach, whereas the other half uses a combined architecture that performs localization and pose estimation simultaneously. Interestingly, the position error distributions are nearly identical, while separate localization significantly outperforms the combined approach in terms of the orientation. This suggests that localization does not bring any benefits in terms of detecting the position, having it predicted simultaneously with the orientation of the satellite is just as accurate. However, the capability to crop irrelevant parts and zoom in on the important part of the image makes a significant difference in orientation estimation. Specifically, the mean orientation error and deviation is $29.66 ^\circ \pm  46.10^\circ$ as opposed to $48.03 ^\circ \pm  49.38^\circ$ of the combined approach. 

\begin{figure*}[!p]
    \centering
    \includegraphics[width=1\textwidth]{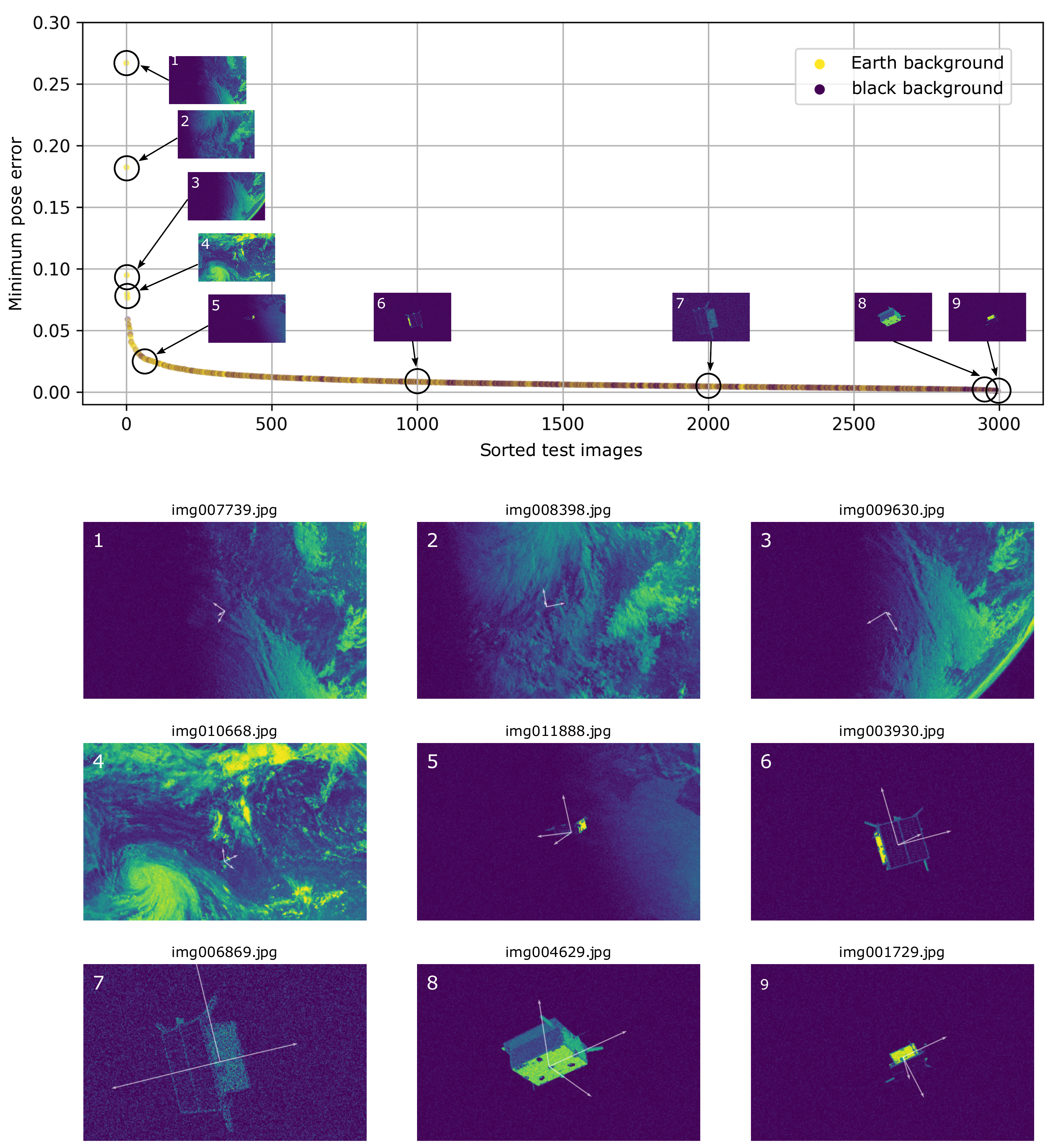}
    \caption{\textit{Top}: Test images ranked by difficulty, measured as minimum pose error across all submissions. \textit{Bottom}: Nine example images from different parts of the distribution. Images are shown with scaled colors to maximize contrast.}
    \label{fig:difficulty}
\end{figure*}

\subsection{Difficulty of samples}

In order to uncover which factors contribute the most to the difficulty of the satellite pose estimation task, the best prediction from all submissions is selected for each image of the test set. This `super pose estimator' is used as a proxy of how difficult the pose estimation task is on a certain sample. The resulting score distribution is plotted in Fig.~\ref{fig:difficulty} along with a number of selected images. Except for a few outliers, the error distribution is flat with pose errors well below $0.05$. In fact, the average orientation error and its standard deviation is $ 0.34^\circ \pm 0.38^\circ$, while the average position error is $0.09 \pm 0.09$ m.\footnote{In comparison, the winning team \textsc{UniAdelaide} achieved $0.41^\circ \pm 1.50^\circ$ orientation error and $0.13 \pm 0.09$ m  relative position error.}

The general trend is that the images with black background, representing the case of an under-exposed star field, are easier compared to the samples with Earth background. Black background makes the detection of the satellite a straightforward task, given the sharp contrast of the satellite to its background. Having a cluttered Earth background makes the pose estimation more difficult.

\begin{figure*}[!t]
    \centering
    \includegraphics[width=1\textwidth]{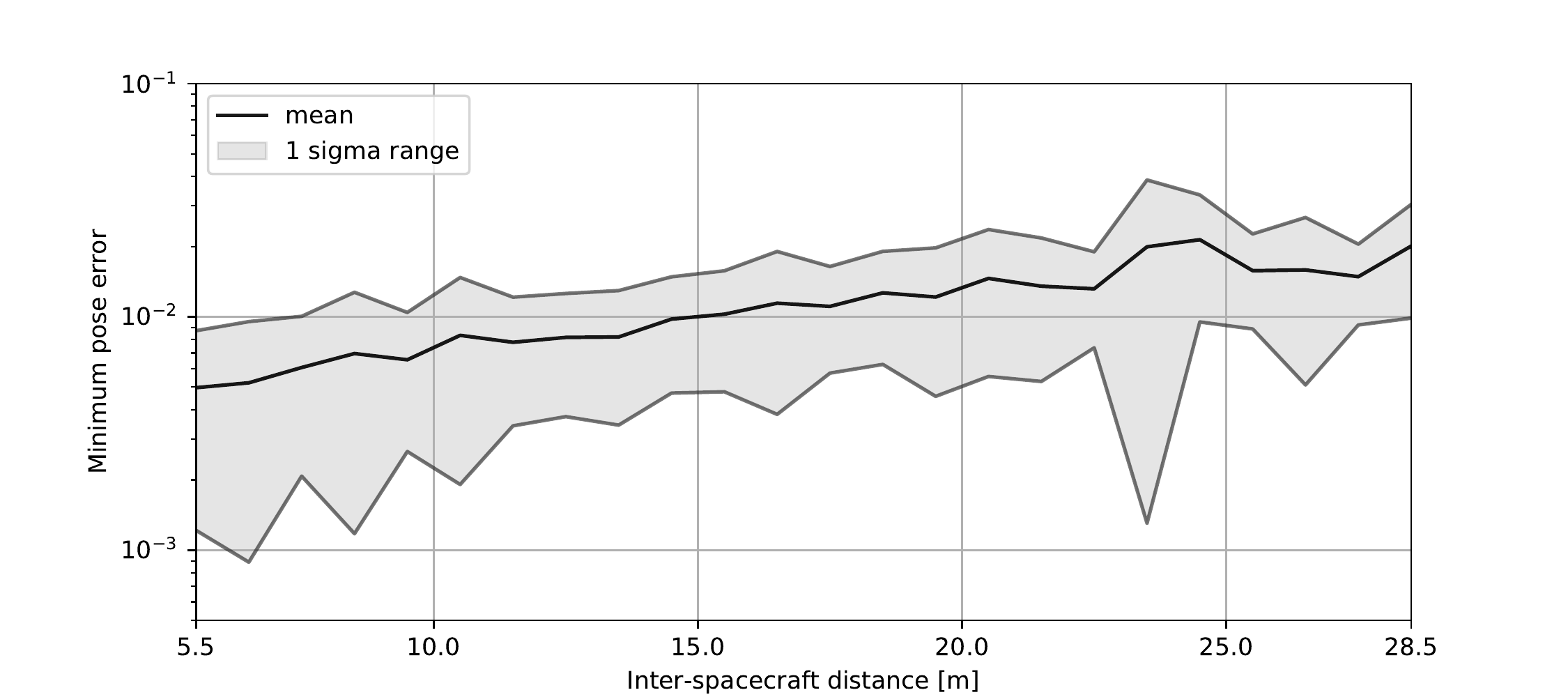}
    \caption{Distribution of the minimum pose error in logarithmic scale with respect to the inter-spacecraft distance. The minimum pose error is computed across all submissions. Mean and standard deviation are calculated over one meter wide distance bins.}
    \label{fig:dist_vs_score}
\end{figure*}

The most challenging samples are the images with Earth background and small target due to large inter-spacecraft distance. In this situation, the apparent size of the satellite can be comparable with features on the background image, and in some cases the contrast of the satellite to the background is minimal. This makes pose estimation particularly challenging. In fact, just spotting the satellite in these images is a demanding task for humans as well (see first four images in Fig.~\ref{fig:difficulty}). Fig.~\ref{fig:dist_vs_score} also highlights the importance of the inter-spacecraft distance. The figure plots the distribution of minimum pose errors with respect to the inter-spacecraft distance, with the mean and standard deviation of minimum pose errors calculated over 1 meter wide distance bins. The distribution of scores is correlated with the target distance, i.e., it is harder to estimate the pose of the satellites that are farther away. This is expected, since larger target distance results in a smaller apparent size of the satellite, corresponding to less pixels associated with the spacecraft.

\section{Conclusion and future work}
\label{sec:conclusion}

The aim of organizing the Satellite Pose Estimation Challenge (SPEC) was to draw more attention to the satellite pose estimation problem and to provide a benchmark to gauge different approaches. Nearly 50 teams participated during the 5 month long duration of the competition. This paper summarizes the creation of the dataset and the considerations put into designing this competition. Based on the submissions and a survey conducted amongst the top performing participants, the analysis is presented on different approaches to the problem. The top performing participants managed to significantly outperform the previous state-of-the-art and push the boundaries of the vision-based satellite pose estimation further.

The analysis on the submissions discovered that the target distance and cluttered backgrounds are the most significant factors contributing to the difficulty of samples. A general trend in computer vision also observed in this competition is the domination of deep learning approaches. Virtually all teams relied on Deep Neural Networks (DNN), at least in some steps of their pose estimation pipeline. However, while DNNs proved to be indispensable in solving the problem of perception, they are still not the best choice throughout all steps of a pose estimation pipeline. Perspective-n-Point (PnP)-based keypoint matching techniques that used keypoints detected by DNNs won the first two places. Another finding was that with the availability of high resolution images and Graphical Processing Unit (GPU) memories that limit input resolution, a separate localization step can bring significant improvements in pose accuracy, as it allows for cropping the irrelevant parts of the image.

Overall, the scores of the top submissions indicate that various DNN architectures are able to perform good pose estimation of a uncooperative spacecraft, provided the servicer has access to the target's 3D model or 3D keypoint coordinates as designed by the mission operators. However, the performances of the same architectures on real images are relatively poor, as the real images have different statistical distributions from the synthetic images that were used to train the DNNs. As any DNNs deployed in future space missions will undoubtedly utilize synthetic images as a main source of training, future SPEC must design the datasets and competition metrics that better reflect the significance of domain gap. Ultimately, to support debris removal and other representative mission scenarios, future SPEC must address the issue of estimating the pose of an unknown resident space object. 

\section*{Acknowledgment}
The authors would like to thank OHB Sweden for the 3D model of the Tango spacecraft used to create the images used in this work and for the flight images collected during the PRISMA extended mission.

\bibliographystyle{IEEEtran}  
\bibliography{references}  

\begin{IEEEbiography}[{\includegraphics[width=1in,height=1.25in,clip,keepaspectratio]{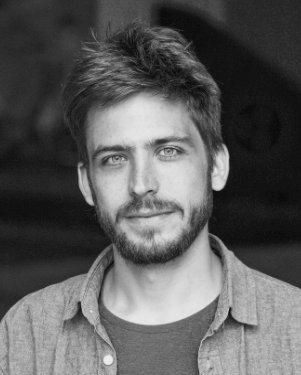}}]{Mate Kisantal} received a B.Sc.~degree in Vehicle Engineering from the Budapest University of Technology and Economics and an M.Sc.~degree from Delft University of Technology, where he did research in the Micro Air Vehicle Laboratory. In 2018 he joined the European Space Agency as a Young Graduate Trainee in Artificial Intelligence, and worked in the Advanced Concepts Team. His research interest is in the intersection of mobile robotics, autonomy, computer vision, and machine learning.
\end{IEEEbiography}

\begin{IEEEbiography}[{\includegraphics[width=1in,height=1.25in,clip,keepaspectratio]{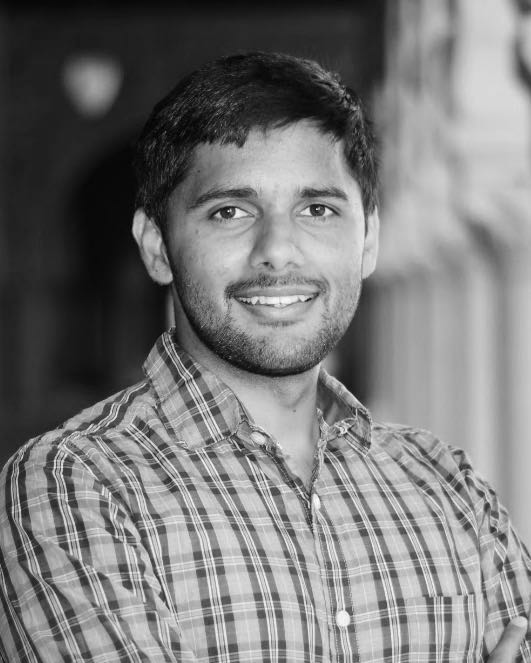}}]{Sumant Sharma} received the B.S.~degree from the Georgia Institute of Technology (2013), M.S.~degree (2015), and Ph.D.~degree (2019) from the Department of Aeronautics and Astronautics at Stanford University. As a Ph.D.~student in the Space Rendezvous Laboratory at Stanford University, his research focused on monocular computer vision algorithms to enable navigation systems for on-orbit servicing and rendezvous missions requiring close proximity. During 2016 and 2017, he was the Co-chairman of the Nominations Commission of the Associated Students of Stanford University, overseeing the appointment of Stanford students to university committees. From 2017 to 2018, he worked as a Systems Engineer at NASA Ames Research Center comparing the performance of machine learning-based methods against conventional feature-based methods for on-orbit servicing applications. Dr.~Sharma is currently a computer vision engineer at Wisk Aero, a company based in Mountain View, California, working on navigation algorithms for electric-powered air transportation. Dr.~Sharma is currently a peer reviewer for the IEEE Transactions on Aerospace and Electronic Systems and IEEE Access.
\end{IEEEbiography}

\begin{IEEEbiography}[{\includegraphics[width=1in,height=1.25in,clip,keepaspectratio]{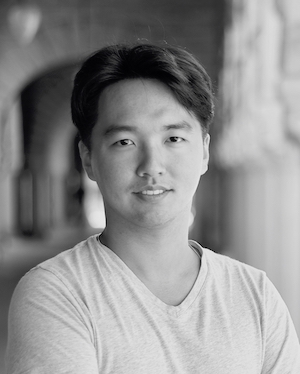}}]{Tae Ha Park} is a Ph.D.~candidate in the Space Rendezvous Laboratory, Stanford University. He graduated from Harvey Mudd College with a Bachelor of Science degree (2017) in engineering. His research interest is in the development of machine learning techniques and GN\&C algorithms for spaceborne computer vision tasks, specifically on robust and accurate determination of the relative position and attitude of arbitrary resident space objects using monocular vision. Potential applications include space debris removal and refueling of defunct geostationary satellites with unprecedented autonomy and safety measures.
\end{IEEEbiography}

\begin{IEEEbiography}[{\includegraphics[width=1in,height=1.25in,clip,keepaspectratio]{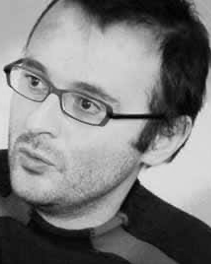}}]{Dario Izzo} graduated as a Doctor of Aeronautical Engineering from the University Sapienza of Rome (Italy). He then took a second master in Satellite Platforms at the University of Cranfield in the United Kingdom and completed his Ph.D.~in Mathematical Modelling at the University Sapienza of Rome where he lectured classical mechanics and space flight mechanics.

Dario Izzo later joined the European Space Agency and became the scientific coordinator of its Advanced Concepts Team. He devised and managed the Global Trajectory Optimization Competitions events, the ESA Summer of Code in Space and the Kelvins innovation and competition platform. He published more than 170 papers in international journals and conferences making key contributions to the understanding of flight mechanics and spacecraft control and pioneering techniques based on evolutionary and machine learning approaches.

Dario Izzo received the Humies Gold Medal and led the team winning the 8$^\textrm{th}$ edition of the Global Trajectory Optimization Competition.
\end{IEEEbiography}

\begin{IEEEbiography}[{\includegraphics[width=1in,height=1.25in,clip,keepaspectratio]{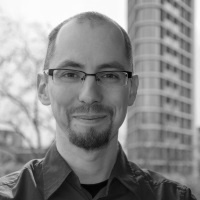}}]{Marcus M\"artens} graduated from the University of Paderborn (Germany) with a Masters degree in computer science. He joined the European Space Agency as a Young Graduate Trainee in artificial intelligence where he worked on multi-objective optimization of spacecraft trajectories. He was part of the winning team of the 8th edition of the Global Trajectory Optimization Competition (GTOC) and received a HUMIES gold medal for developing algorithms achieving human competitive results in trajectory design. The Delft University of Technology awarded him a Ph.D. for his thesis on information propagation in complex networks. After his time at the network architectures and services group in Delft (Netherlands), Marcus rejoined the European Space Agency, where he works as a research follow in the Advanced Concepts Team. While his main focus is on applied artificial intelligence and evolutionary optimization, Marcus has worked together with experts from different fields and authored works related to neuroscience, cyber-security and gaming.\end{IEEEbiography}

\begin{IEEEbiography}[{\includegraphics[width=1in,height=1.25in,clip,keepaspectratio]{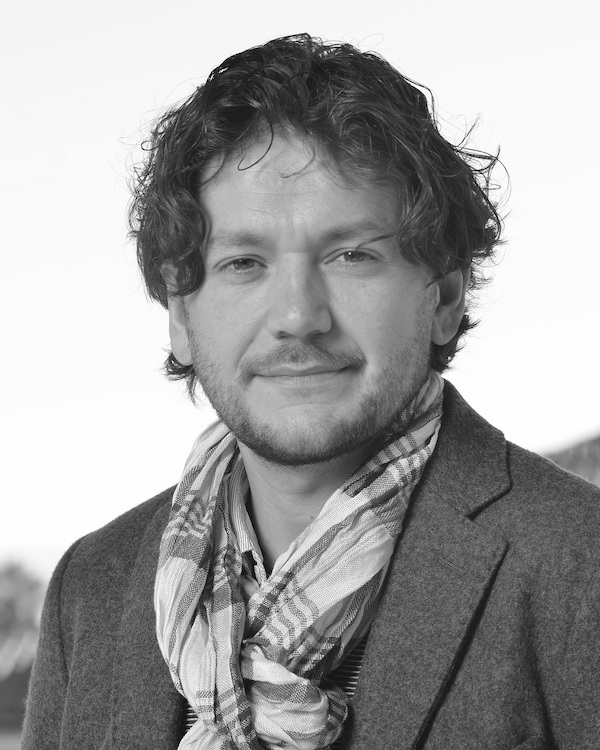}}]{Simone D'Amico} received the B.S.~and M.S.~degrees from Politecnico di Milano (2003) and the Ph.D.~degree from Delft University of Technology (2010). From 2003 to 2014, he was research scientist and team leader at the German Aerospace Center (DLR). There, he gave key contributions to the design, development, and operations of spacecraft formation-flying and rendezvous missions such as GRACE (United States/Germany), TanDEM-X (Germany), PRISMA (Sweden/Germany/France), and PROBA-3 (ESA). Since 2014, he has been Assistant Professor of Aeronautics and Astronautics at Stanford University, Founding director of the Space Rendezvous Laboratory (SLAB), and Satellite Advisor of the Student Space Initiative (SSSI), Stanford’s largest undergraduate organization. He has over 150 scientific publications and 2500 google scholar’s citations, including conference proceedings, peer-reviewed journal articles, and book chapters. D'Amico's research aims at enabling future miniature distributed space systems for unprecedented science and exploration. His efforts lie at the intersection of advanced astrodynamics, GN\&C, and space system engineering to meet the tight requirements posed by these novel space architectures. The most recent mission concepts developed by Dr.~D'Amico are a miniaturized distributed occulter/telescope (mDOT) system for direct imaging of exozodiacal dust and exoplanets and the Autonomous Nanosatellite Swarming (ANS) mission for characterization of small celestial bodies. He is Chairman of the NASA's Starshade Science and Technology Working Group (TSWG) and Fellow of the NAE’s US FOE Symposium. D’Amico’s research is supported by NASA, NSF, AFRL, AFOSR, KACST, and Industry. He is member of the advisory board of space startup companies and VC edge funds. He is member of the Space-Flight Mechanics Technical Committee of the AAS, Associate Fellow of AIAA, Associate Editor of the AIAA Journal of Guidance, Control, and Dynamics and the IEEE Transactions of Aerospace and Electronic Systems. Dr.~D’Amico was recipient of the Leonardo 500 Award by the Leonardo Da Vinci Society and ISSNAF (2019), the Stanford’s Introductory Seminar Excellence Award (2019), the FAI/NAA‘s Group Diploma of Honor (2018), the Exemplary System Engineering Doctoral Dissertation Award by the International Honor Society for Systems Engineering OAA (2016), the DLR’s Sabbatical/Forschungssemester in honor of scientific achievements (2012), the DLR’s Wissenschaft Preis in honor of scientific achievements (2006), and the NASA’s Group Achievement Award for the Gravity Recovery and Climate Experiment, GRACE (2004).
\end{IEEEbiography}


\end{document}